%% file: main.tex
\documentclass{article}

\usepackage[final, nonatbib, preprint]{neurips_2022}

\usepackage[utf8]{inputenc} % allow utf-8 input
\usepackage[T1]{fontenc}    % use 8-bit T1 fonts
\usepackage{url}            % simple URL typesetting
\usepackage{booktabs}       % professional-quality tables
\usepackage{amsfonts}       % blackboard math symbols
\usepackage{nicefrac}       % compact symbols for 1/2, etc.
\usepackage{microtype}      % microtypography
\usepackage{xfrac}
\usepackage{enumitem}
\usepackage{amsmath}
\usepackage{mathrsfs}
\usepackage{amsthm}

\newtheorem{theorem}{Theorem}

\newtheorem{proposition}{Proposition}
\newtheorem{assumption}[theorem]{Assumption}

\newtheorem{innercustomgeneric}{\customgenericname}
\providecommand{\customgenericname}{}
\newcommand{\newcustomtheorem}[2]{%
  \newenvironment{#1}[1]
  {%
   \renewcommand\customgenericname{#2}%
   \renewcommand\theinnercustomgeneric{##1}%
   \innercustomgeneric
  }
  {\endinnercustomgeneric}
}
\newcustomtheorem{customprop}{Proposition}

\usepackage[dvipsnames]{xcolor}
\usepackage{amsmath}
\usepackage{amsfonts}
\usepackage{amssymb}
\usepackage{bm,epsf}

\usepackage{tcolorbox}

\usepackage{comment}

\usepackage{fancyhdr}
\usepackage{graphicx}
\usepackage{subcaption}

\numberwithin{equation}{section}

\usepackage{hyperref}
\hypersetup{
    colorlinks=true,
    linkcolor=blue,
    filecolor=magenta,      
    urlcolor=NavyBlue,
    citecolor=Blue
}

\usepackage{chngcntr}

\newcommand{\lRo}{\left(}
\newcommand{\rRo}{\right)}
\newcommand{\lSq}{\left[}
\newcommand{\rSq}{\right]}
\newcommand{\lCu}{\left\lbrace}
\newcommand{\rCu}{\right\rbrace}

\newcommand{\Nc}{\mathcal{N}}

\newcommand{\norm}[1]{\left\lVert#1\right\rVert}
\newcommand{\expec}[1]{\mathbb{E}\left[#1\right]}

\newcommand{\sphere}[1]{\mathbb{S}^{#1-1}}

\DeclareMathOperator*{\argmin}{arg\,min}

\newcommand{\RR}{\bm{R}}

\newcommand{\R}{\mathbb{R}}

\newcommand{\fst}{1\textsuperscript{st}}
\newcommand{\snd}{2\textsuperscript{nd}}
\newcommand{\trd}{3\textsuperscript{rd}}

\title{Learning sparse features can lead to overfitting in neural networks}

\author{
    Leonardo Petrini \thanks{Equal contribution (a coin was flipped).} \\
    Institute of Physics\\
    \'Ecole Polytechnique F\'ed\'erale de Lausanne\\
    \texttt{leonardo.petrini@epfl.ch} \\
    \And
    Francesco Cagnetta \footnotemark[1] \\
    Institute of Physics\\
    \'Ecole Polytechnique F\'ed\'erale de Lausanne\\
    \texttt{francesco.cagnetta@epfl.ch} \\
  \And
  Eric Vanden-Eijnden \\
  Courant Institute of Mathematical Sciences\\
  New York University\\
  \texttt{eve2@cims.nyu.edu}
  \And
    Matthieu Wyart \\
    Institute of Physics\\
    \'Ecole Polytechnique F\'ed\'erale de Lausanne\\
    \texttt{matthieu.wyart@epfl.ch} \\
}

\begin{document}
	
	\maketitle

\begin{abstract}
It is widely believed that the success of deep networks lies in their ability to learn a meaningful representation of the features of the data. Yet, understanding  when and how this feature learning improves performance remains a challenge: for example, it is beneficial for modern architectures trained to classify images, whereas it is detrimental for fully-connected networks trained on the same data. Here we propose an explanation for this puzzle, by showing that feature learning can perform worse than lazy training (via random feature kernel or the NTK) as the former can lead to a sparser neural representation. Although sparsity is known to be essential for learning anisotropic data, it is detrimental when the target function is constant or smooth along certain directions of input space. We illustrate this phenomenon in two settings: \emph{(i)} regression of Gaussian random functions on the $d$-dimensional unit sphere and  \emph{(ii)} classification of benchmark datasets of images. For \emph{(i)}, we compute the scaling of the generalization error with the number of training points and show that methods that do not learn features generalize better, even when the dimension of the input space is large. For \emph{(ii)}, we show empirically that learning features can indeed lead to sparse and thereby less smooth representations of the image predictors. This fact is plausibly responsible for deteriorating the performance, which is known to be correlated  with smoothness along diffeomorphisms.
\end{abstract}

\section{Introduction}

Neural networks are responsible for a technological revolution in a variety of machine learning tasks. Many such tasks require learning functions of high-dimensional inputs from a finite set of examples, thus should be generically hard due to the \textit{curse of dimensionality} \cite{luxburg2004distance,bach2017breaking}: the exponent that controls the scaling of the generalization error with the number of training examples is inversely proportional to the input dimension $d$. For instance, for standard image classification tasks with $d$ ranging in $10^3\div 10^5$, such exponent should be practically vanishing, contrary to what is observed in practice~\cite{hestness2017deep}. In this respect, understanding the success of neural networks is still an open question. A popular explanation is that, during training, neurons adapt to features in the data that are relevant for the task \cite{le2013building}, effectively reducing the input dimension and making the problem tractable \cite{shwartz2017opening,ansuini2019intrinsic,recanatesi2019dimensionality}. However, understanding quantitatively if this intuition is true and how it depends on the structure of the task remains a challenge. 

Recently much progress was made in characterizing the conditions which lead to features learning, in the overparameterized setting where networks generally perform best. When the initialization scale of the network parameters is large \cite{chizat2019lazy} one encounters the \textit{lazy training regime}, where neural networks  behave as  kernel methods \cite{jacot2018neural, Du2019} (coined Neural Tangent Kernel or NTK) and features are not learned. By contrast, when the initialization scale is small, a \textit{feature learning regime} is found \cite{rotskoff2018neural,mei2018mean,sirignano2018mean} where the network parameters evolve significantly during training. This limit is much less understood apart from very simple architectures, where it can be shown to lead to sparse representations where a limited number of neurons are active after training \cite{woodworth2020kernel}. Such sparse representations can also be obtained by regularizing the weights during training \cite{bach2017breaking,de2020sparsity}.

\begin{figure}
    \centering
    \includegraphics[width=\linewidth]{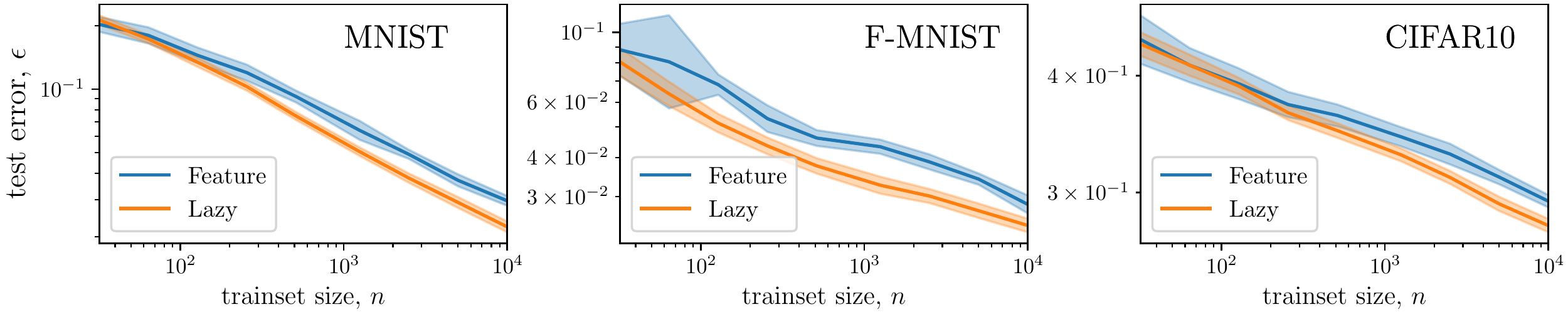}
    \caption{\textbf{Feature vs. Lazy in image classification.} Generalization error as a function of the training-set size $n$ for infinite-width fully-connected networks (FCNs) trained in the feature (blue) and lazy regime (orange). In the latter case the limit is taken exactly by training an SVC algorithm with the analytical NTK \cite{chen_equivalence_2021}. In the former case, 
    the infinite-width limit can be accurately approximated for these datasets by considering very wide nets ($H = 10^3$), and performing ensemble averaging on different initial conditions of the parameters as shown in \cite{geiger2019scaling, geiger_landscape_2021}. Panels correspond to different benchmark image datasets \cite{lecun1998gradient, xiao_fashion-mnist_2017, krizhevsky_learning_2009}. Results are averaged over 10 different initializations of the networks and datasets.}
    \label{fig:learning_curves_images}
\end{figure}

In terms of performance, most theoretical works have focused on fully-connected networks. For these architectures, feature learning  was shown to significantly outperform lazy training~\cite{chizat2020implicit,ghorbani2020neural,refinetti2021classifying,paccolat2020compressing,rotskoff2018neural} for certain tasks, including approximating a function which depends only on a subset or a linear combination of the input variables. However, when such primitive networks are trained on image datasets, learning features is detrimental \cite{geiger2019disentangling,lee2020finite}, as illustrated in \autoref{fig:learning_curves_images} (see ~\cite[Fig.~3]{paccolat2020compressing} for the analogous plot in the case of a target function depending on just one of the input variables, where learning features is beneficial). A similar result was observed in simple models of data \cite{ortiz2021can}.  These facts are unexplained, yet central to understanding the implicit bias of the feature learning regime.

\subsection{Our contribution}

Our main contribution is to provide an account of the drawbacks of learning sparse representations based on the following set of ideas. Consider, for concreteness, an image classification problem: \textit{(i)} images class varies little along smooth deformations of the image; \textit{(ii)} due to that, tasks like image classification require a continuous distribution of neurons to be represented; \textit{(iii)} thus, requiring sparsity can be detrimental for performance. We build our argument as follows.

\begin{itemize}

\item In order to find a quantitative description of the phenomenon, we start from the problem of regression of a random target function of controlled smoothness on the $d$-dimensional unit sphere, and study the property of the minimizers of the empirical loss with $n$ observations, both in the lazy and the feature learning regimes. More specifically, we consider two extreme limits---the NTK limit and mean-field limit---as representatives of lazy and feature regimes, respectively (\autoref{sec:problem}). Both these limits admit a simple formulation that allows us to predict generalization performances. In particular, our results on feature learning rely on solutions having an atomic support. This property can be justified for one-hidden-layer neural networks with ReLU activations and weight decay. Yet, we also find such a sparsity empirically using gradient descent in the absence of regularization, if weights are initialized to be small enough.

\item We find that lazy training leads to smoother predictors than feature learning.  As a result, lazy training outperforms feature learning when the target function is also sufficiently smooth. Otherwise, the performances of the two methods are comparable, in the sense that they display the same asymptotic decay of generalization error with the number of training examples. Our predictions are obtained from  asymptotic arguments that we systematically back up with numerical studies.
    
    \item For image datasets, it is believed that diffeomorphisms of images are key transformations along which the predictor function should only mildly vary to obtain good performance \cite{bruna2013invariant}. From the results above, a natural explanation as to why lazy beats feature for fully connected networks is that it leads to predictors with smaller variations along diffeomorphisms. We confirm that this is indeed the case empirically on benchmark datasets.

\end{itemize}

Numerical experiments are performed in PyTorch \cite{paszke_pytorch_2019}, and the code for reproducing experiments is available online at
\href{https://github.com/pcsl-epfl/regressionsphere}{github.com/pcsl-epfl/regressionsphere}.

\subsection{Related Work}

The property that training ReLU networks in the feature regime leads to a sparse representation was observed empirically \cite{maennel2018gradient}. This property can be justified for one-hidden-layer networks by casting training as a L1 minimization problem~\cite{neyshabur_norm-based_2015, bach2017breaking}, then using a representer theorem~\cite{boyer2019representer,de2020sparsity, chizat2021sparse}. This is analogous to what is commonly done in predictive sparse coding \cite{olshausen_emergence_1996, mairal_supervised_2008, mehta_sparsity-based_nodate, sulam_adversarial_2021}.

Many works have investigated the benefits of learning sparse representations in neural networks. \cite{bach2017breaking, chizat2020implicit,ghorbani2020neural,refinetti2021classifying, paccolat2020compressing, yehudai2019power,ghorbani2019limitations} study cases in which the true function only depends on a linear subspace of input space, and show that feature learning profitably capture such property. Even for more general problems, sparse representations of the data might emerge naturally during deep network training---a phenomenon coined ~\textit{neural collapse} \cite{papyan2020prevalence}. Similar sparsification phenomena, for instance, have been found to allow for learning convolutional layers from scratch~\cite{neyshabur_towards_2020, ingrosso2022data}. Our work builds on this body of literature by pointing out that learning sparse features can be detrimental, if the task does not allow for it.

There is currently no general framework to predict rigorously the learning curve exponent $\beta$ defined as $\epsilon(n)={\cal O}( n^{-\beta})$ for kernels. Some of our asymptotic arguments can be obtained by other approximations, such as assuming that data points lie on a lattice in $\mathbb{R}^d$~\cite{spigler2019asymptotic}, or by using the non-rigorous replica method of statistical physics~\cite{bordelon2020spectrum,cui_generalization_2021,tomasini2022failure}. In the case $d=2$, we provide a more explicit mathematical formulation of our results, which leads to analytical results for certain kernels. We systematically back up our predictions with numerical tests as $d$ varies.

Finally, in the context of image classification, the connection between performance and `stability'  or smoothness  toward small diffeomorphisms of the inputs has been conjectured by \cite{bruna2013invariant, mallat2016understanding}. Empirically, a strong correlation between these two quantities was shown to hold across various architectures for real datasets~\cite{petrini_relative_2021}. In that reference, it was found that fully connected networks lose their stability over training: here we show that this effect is much less pronounced in the lazy regime.

\section{Problem and notation}\label{sec:problem}

\begin{figure}
    \centering
    \hspace*{-.5cm}
    \includegraphics[width=.87\textwidth]{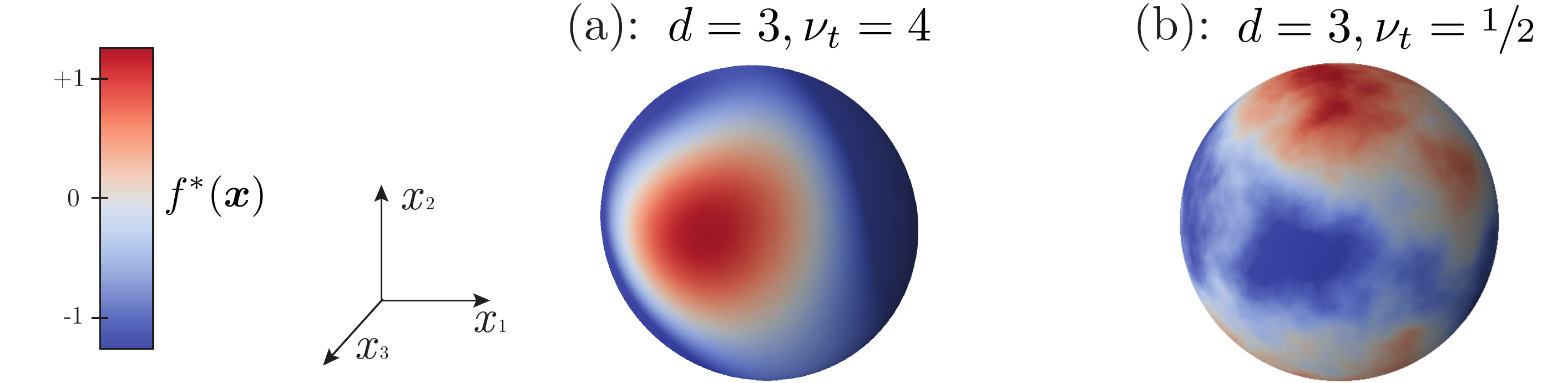}
    \caption{\textbf{Gaussian random process on the sphere.} We show here two samples of the task introduced in \autoref{sec:problem} when the target function $f^*(\bm x)$ is defined on the $3-$dimensional unit sphere. (a) and (b) show samples of large and small smoothness coefficient $\nu_t$, respectively.}
    \label{fig:spherical_data}
\end{figure}

\paragraph{Task} We consider a supervised learning scenario with $n$ training points $\{\bm{x}_i\}_{i=1}^n$ uniformly drawn on the $d$-dimensional unit sphere $\sphere{d}$. We assume that the target function $f^*$ is an isotropic Gaussian random process on $\sphere{d}$ and control its statistics via the spectrum: by introducing the decomposition of $f^*$ into spherical harmonics~(see~\autoref{app:spherical} for definitions),
\begin{equation}\label{eq:target}
    f^*(\bm{x}) = \sum_{k\geq 0}\sum_{\ell=1}^{\Nc_{k,d}} f^*_{k,\ell}Y_{k,\ell}(\bm{x})\quad \text{with} \quad  \expec{f^*_{k,\ell}}=0,\quad \expec{f^*_{k,\ell}f^*_{k',\ell'}} = c_k\delta_{k,k'}\delta_{\ell,\ell'}.
\end{equation}
We assume that all the $c_{k}$ with $k$ odd vanish apart from $c_1$: this is required to guarantee that $f^*$ can be approximated as well as desired with a one-hidden-layer ReLU network with no biases, as discussed in~\autoref{app:spherical}. We also assume that the non-zero $c_k$ decay as a power of $k$ for $k\gg 1$, $c_k\sim k^{-2\nu_t -(d-1)}$. The exponent $\nu_t\,{>}\,0$ controls the (weak) differentiability of $f^*$ on the sphere (see~\autoref{app:spherical}) and also the statistics of $f^*$ in real space:
\begin{equation}\label{eq:diff-via-kernel}
\expec{\vert f^*(\bm{x})-f^*(\bm{y})\rvert^2}=O\lRo |\bm{x}-\bm{y}|^{2\nu_t} \rRo =O\lRo \lRo1-\bm{x}\cdot\bm{y} \rRo^{\nu_t} \rRo \quad\text{ as }\ \ \bm{x}\to\bm{y}.\end{equation}
Examples of such a target function for $d=3$ and different values of $\nu_t$ are reported in \autoref{fig:spherical_data}.

\paragraph{Neural network representation in the feature regime} In this regime we aim to approximate  the target function $f^*(x)$ via a \emph{one-hidden-layer neural network} of width $H$,
\begin{equation}\label{eq:model}
f_H(\bm{x}) = \frac{1}{H} \sum_{h=1}^H w_h \sigma( \bm{\theta}_h\cdot\bm{x}),
\end{equation}
where $\lCu \bm{\theta}_h \rCu_{h=1}^H$ (the features) and $\lCu w_h \rCu_{h=1}^H$ (the weights) are the network parameters to be optimized, and $\sigma(x)$ denotes the ReLU function, $\sigma(x)\,{=}\,\text{max}\lCu0,x\rCu$. 
If we assume that $\{\bm{\theta}_h,w_h\}_{h=1}^H$ are independently drawn from a probability measure $\mu$ on $\sphere{d}\times \mathbb{R}$ such that the Radon measure $\gamma = \int_{\R} w \mu(\cdot,dw)$ exists, then as $H\to\infty$,
\begin{equation}
    \label{eq:MF}
    \lim_{H\to\infty} f_H(\bm{x}) = \int_{\sphere{d}} \sigma(\bm{\theta}\cdot\bm{x}) d\gamma(\bm{\theta}) \qquad \text{a.e. on $\sphere{d}$}.
\end{equation}
This is the so-called mean-field limit~\cite{rotskoff2018neural, mei2018mean}, and it is then natural to determine the optimal $\gamma$ via
\begin{equation}
       \gamma^*= \argmin_{\gamma} \int_{\sphere{d}} \lvert d\gamma(\bm{\theta})\rvert \quad \text{subject to:} \quad  \int_{\sphere{d}} \,\sigma(\bm{\theta}\cdot\bm{x}_i)d\gamma(\bm{\theta}){=}f^*(\bm{x}_i)\quad \forall i=1,\dots,n. 
       \label{eq:predictor-feature}
\end{equation}
In practice, we can approximate this minimization problem by using a network with large but finite width, constraining the feature to be on the sphere $|\bm{\theta}_h|=1$, and minimizing the following empirical loss with L1 regularization on the weights,
\begin{equation}\label{eq:predictor:FR1}
       \min_{\substack{\{w_h,\theta_h\}_{h=1}^H \\ |\bm{\theta}_h|=1}}  \frac{1}{2n}\sum_{i=1}^n\left( f^*(\bm{x}_i)-\frac{1}{H} \sum_{h=1}^H w_h \sigma( \bm{\theta}_h\cdot\bm{x}_i) \right)^2  + \frac{\lambda}{ H}\sum_{h=1}^H|w_h|.
\end{equation}

This minimization problem leads to~\eqref{eq:predictor-feature} when $H\to\infty$ and $\lambda \to0$. Note that, by homogeneity of ReLU, \eqref{eq:predictor:FR1} can be shown to be equivalent to imposing a regularization on the L2 norm of all parameters~\cite[Thm. 10]{neyshabur_norm-based_2015}, i.e. the usual weight decay.

To proceed we will make the following assumption about the minimizer $\gamma^*$:
\begin{assumption}
\label{as:sparse:AR}
The minimizer $\gamma^*$ of \eqref{eq:predictor-feature} is unique and atomic, with $n_A \le n$ atoms, i.e. there exists $\{w^*_i,\bm{\theta}^*_i\}_{i=1}^{n_A}$ such that
\begin{equation}
    \label{eq:N:atoms}
    \gamma^* = \sum_{i=1}^{n_A} w^*_i \delta_{\bm{\theta}^*_i}.
\end{equation}
\end{assumption}
The main component of the assumption is the uniqueness of $\gamma^*$; if it holds the sparsity of $\gamma^*$ follows from the representer theorem, see e.g.~\cite{boyer2019representer}. Both the uniqueness and sparsity of the minimizer can be justified as holding generically using asymptotic arguments involving recasting the $L1$ minimization problem~\ref{eq:predictor-feature} as a linear programming one: these arguments are standard (see e.g.~\cite{chen1998atomic}) and are presented in \autoref{app:l1minimization} for the reader convenience. In our arguments below to deduce the scaling of the generalization error we will mainly use that $n_A=O(n)$---we shall confirm this fact numerically even in the absence of regularization, if the weights are initialized to be small enough. Notice that from Assumption \ref{as:sparse:AR} it follows that the predictor in the feature regime corresponding to the minimizer $\gamma^*$ takes the following form
\begin{equation}
\label{eq:predictor-feature-assumption1}
    f^{\text{FEATURE}}(\bm x) = \sum_{i=1}^{n_A} w_i^* \sigma(\bm{\theta^*}_i\cdot \bm{x}).
\end{equation}

\paragraph{Neural network representation in the lazy regime.} In this regime we approximate the target function $f^*(x)$ via 
\begin{equation}\label{eq:sol-lazy}
    f^{\text{NTK}}(\bm{x}) =\sum_{i=1}^n g_i K^{\text{NTK}}(\bm{x}_i\cdot \bm{x}), %\quad\text{ with }     f^*(\bm{x}_j) =\sum_{i=1}^n g_i  K^{\text{NTK}}(\bm{x}_i\cdot \bm{x}_j), \quad j=1,\ldots,n.
\end{equation}
where the weights $\{g_i\}_{i=1}^n$ solve
\begin{equation}\label{eq:eq-lazy}
    f^*(\bm{x}_j) =\sum_{i=1}^n g_i  K^{\text{NTK}}(\bm{x}_i\cdot \bm{x}_j), \qquad j=1,\ldots,n.
\end{equation}
and $K^{\text{NTK}}(\bm{x}\cdot \bm{y})$ is the \emph{Neural Tangent Kernel} (NTK)~\cite{jacot2018neural} 
\begin{equation}\label{eq:ntk-kernel}
        K^{\text{NTK}}(\bm{x}\cdot \bm{y}) = \int_{\sphere{d}\times\mathbb{R}} \left( \sigma(\bm{\theta}\cdot\bm{x})\sigma(\bm{\theta}\cdot\bm{y}) + w^2 \, \bm{x}\cdot \bm{y}\, \sigma'(\bm{\theta}\cdot\bm{x}) \sigma'(\bm{\theta}\cdot\bm{y})\right) d\mu_0(\bm{\theta},w).
\end{equation}
Here $\mu_0$ is a fixed probability distribution which, in the NTK training regime~\cite{jacot2018neural}, is the distribution  of the features and weights at initialization. It is well-known~\cite{scholkopf2001learning} that the solution to kernel ridge regression problem \label{eq:eq-lazy} can also be expressed via the kernel trick as
\begin{equation}
\label{eq:measure-convergence:ntk}
    f^{\text{NTK}}(\bm{x})  = \int_{\sphere{d}\times\R}\!\!\!\! \left(g_w(\bm{\theta},w)\sigma(\bm{\theta}\cdot\bm{x}) + 
    w\bm{x} \cdot \bm{g}_\theta(\bm{\theta},w)\sigma'(\bm{\theta}\cdot\bm{x})\right) d\mu_0(\bm{\theta},w)
\end{equation}
where $\bm{g}_\theta$ and $g_w$  are the solutions of
\begin{equation}
\label{eq:predictor-lazy-ntk}
\begin{aligned}
        & \min_{g_w, \bm{g}_\theta  }\int_{\sphere{d}\times\mathbb{R}}  \left( g^2_w(w,\bm{\theta}) + |\bm{g}_\theta(w,\bm{\theta})|^2\right) d\mu_0(\bm{\theta},w)\\
        \text{subject to:} & \ \int_{\sphere{d}\times\mathbb{R}} \left( g_w(w,\bm{\theta}) \sigma(\bm{\theta}\cdot\bm{x}_i) +w\bm{x}_i\cdot \bm{g}_\theta(w,\bm{\theta})\sigma'(\bm{\theta}\cdot
        \bm{x}_i)\right)d\mu_0(\bm{\theta},w)=f^*(\bm{x}_i)\\
        & \qquad\qquad\qquad\forall i=1,\dots,n.
\end{aligned}
\end{equation}

Another lazy limit can be obtained equivalently by training only the weights while keeping the features to their initialization value. This is equivalent to forcing $\bm{g}_\theta(\bm{\theta},w)$ to vanish in~\autoref{eq:predictor-lazy-ntk}, resulting again in a kernel method. The kernel, in this case, is called \emph{Random Feature Kernel} ($K^{\text{RFK}}$), and can be obtained from~\autoref{eq:ntk-kernel} by setting $d\mu_0(\bm{\theta},w) = \delta_{w=0} d\tilde\mu_0(\bm{\theta})$. The minimizer can then be written as in~\autoref{eq:sol-lazy} with $K^{\text{NTK}}$ replaced by $K^{\text{RFK}}$.

\section{Asymptotic analysis of generalization}

In this section, we characterize the asymptotic decay of the generalization error $\overline{\epsilon}(n)$ averaged over several realizations of the target function $f^*$. Denoting with $d\tau^{d-1}(\bm x)$ the uniform measure on $\sphere{d}$, 
\begin{equation}\label{eq:test-real}
    \overline{\epsilon}(n) = \mathbb{E}_{f^*}\lSq \int d\tau^{d-1}(\bm{x})\, \lRo f^n(\bm{x})-f^*(\bm{x})\rRo^2\rSq = \mathcal{A}_d n^{-\beta} + o(n^{-\beta}),
\end{equation}
for some constant $\mathcal{A}_d$ which might depend on $d$ but not on $n$. Both for the lazy (see \autoref{eq:sol-lazy}) and feature regimes (see \autoref{eq:predictor-feature-assumption1}) the predictor can be written as a sum of $\mathcal O(n)$ terms:
\begin{equation}\label{eq:predictor-general}
    f^n(\bm{x}) = \sum_{j=1}^{\mathcal {O}(n)} g_j \varphi(\bm{x}\cdot\bm{y}_j) := \int_{\sphere{d}} g^n(\bm{y}) \varphi(\bm{x}\cdot\bm{y}) d\tau(\bm{y}).
\end{equation}
In the feature regime, the $g_j$'s ($\bm y_j$) coincide with the optimal weights $w^*_j$ (features $\bm \theta^*_j$), $\varphi$ with the activation function $\sigma$. In the lazy regime, the $\bm y_j$ are the training points $\bm x_j$, $\varphi$ is the neural tangent or random feature kernel the $g_j$'s are the weights solving \autoref{eq:eq-lazy}.
We have defined the density $g^n(\bm{x})=\sum_j |\sphere{d}|g_j \delta(\bm{x}-\bm{y}_j)$ so as to cast the predictor as a convolution on the sphere. Therefore, the projections of $f^n$ onto spherical harmonics $Y_{k,\ell}$ read $f^n_{k,\ell} = g^n_{k,\ell} \varphi_k$, where $g^n_{k,\ell}$ is the projection of $g^n(\bm{x})$ and $\varphi_k$ that of $\varphi(\bm{x}\cdot{\bm{y}})$. %with with $ g^n_{k,l}= \sum_{j} g_j Y_{k,\ell}(\bm{y}_j)$ and  $\tile{\varphi}_k = \int_{\mathbb{S}^{d-1}} \varphi(\bm{x}\cdot\bm{y}) Y_{k,\ell}(\bm{y})\,d\tau(\bm{y})$.
For ReLU neurons one has (as shown in~\autoref{app:spherical})
\begin{equation}\label{eq:unit-decay}
\varphi^{\text{LAZY}}_k \sim k^{-(d-1)-2\nu}\quad\text{ with }\nu=1/2\text{ (NTK)},3/2\text{ (RFK)},\quad    \varphi^{\text{FEATURE}}_k \sim k^{-\frac{d-1}{2}-3/2}. 
\end{equation}

\paragraph{Main Result} Consider a target function $f^*$ with smoothness exponent $\nu_t$ as defined above, with data lying on $\sphere{d}$. If $f^*$ is learnt with a one-hidden-layer network with ReLU neurons in the regimes specified above, then the generalization error follows $\overline{\epsilon}(n)\sim n^{-\beta}$ with:
\begin{subequations}\label{mainr}
\begin{align}
\label{mainr1}
&\beta^{\text{LAZY}} = \frac{\min\lCu 2(d-1) + 4\nu, 2\nu_t \rCu}{d-1}\,\text{ with }\, \nu=\left\lbrace\begin{aligned}& 1/2 \text{ for NTK}, \\ & 3/2 \text{ for RFK}, \end{aligned}\right.,\\
\label{mainr2}
&\beta^{\text{FEATURE}} = \frac{\min\lCu (d-1) + 3, 2\nu_t \rCu}{d-1}.
\end{align}
\end{subequations}
This is our central result. It implies that if the target function is a smooth isotropic Gaussian field (realized for large $\nu_t$), then lazy beats feature, in the sense that training the network in the lazy regime leads to a better scaling of the generalization performance with the number of training points.

\paragraph{Strategy} There is no general framework for a rigorous derivation of the generalization error in the ridgeless limit $\lambda\to 0$: predictions such as that of~\autoref{mainr} can be obtained by either assuming that training points (for \autoref{mainr1}) and neurons (for \autoref{mainr2}) lie on a periodic lattice~\cite{spigler2019asymptotic}, or (for \autoref{mainr1}) using the replica method from physics \cite{bordelon2020spectrum} as shown in~\autoref{app:replica}. Here we follow a different route, by first characterizing the form of the predictor for $d\,{=}\,2$ (proof in~\autoref{app:2drig}). This property alone allows us to determine the asymptotic scaling of the generalization error. We use it to analytically obtain the generalization error in the NTK case 
with a slightly simplified function $\varphi$ (details in~\autoref{app:2dex}). This calculation motivates a simple ansatz for the form of $g^n(\bm{x})$ entering \autoref{eq:predictor-general} and its projections onto spherical harmonics, 
which extends naturally to arbitrary dimension. We confirm the predictions resulting from this ansatz systematically in numerical experiments.
\begin{figure}
    \centering
    \includegraphics[width=\linewidth]{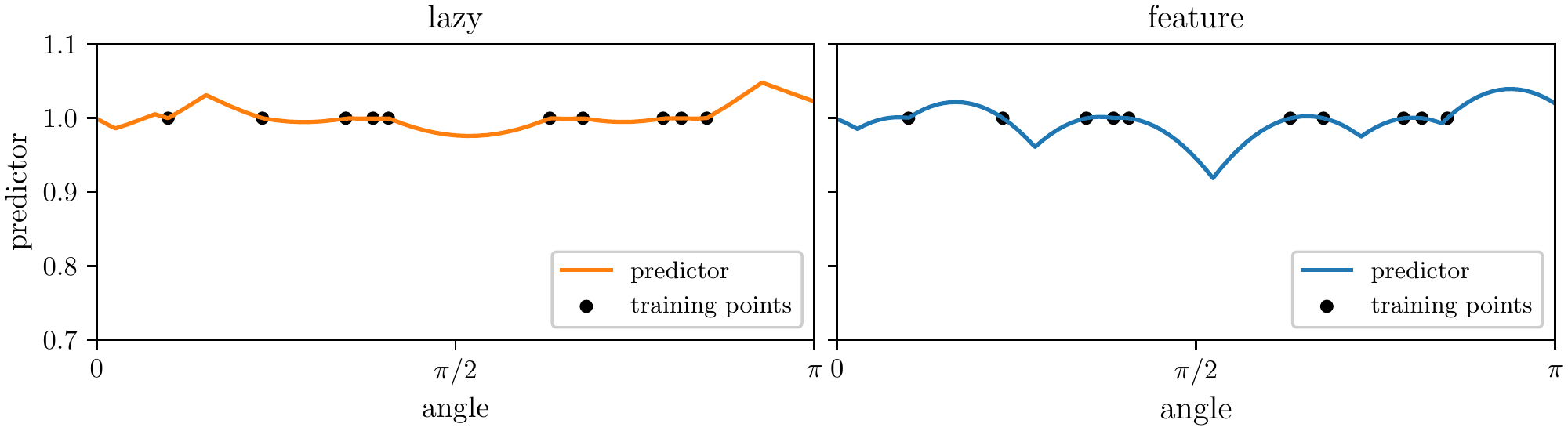}
    \caption{\textbf{Feature vs. Lazy Predictor.} Predictor of the lazy (left) and feature (right) regime when learning the constant function on the ring with $8$ uniformly-sampled training points.}
    \label{fig:parabolas}
\end{figure}

\paragraph{Properties of the predictor in $d=2$}
On the unit circle $\mathbb{S}^1$ all points are identified by a polar angle $x\in[0,2\pi)$. Hence both target function and estimated predictor are functions of the angle, and all functions of a scalar product are in fact functions of the difference in angle. In particular, introducing $\tilde{\varphi}(x)={\varphi}(\cos(x))$,
\begin{equation}\label{eq:predictor-general-2d}
    \begin{aligned}
    f^n(x) = \sum_j g_j \tilde{\varphi}(x-x_j) \equiv \int_0^{2\pi} \frac{dy}{2\pi}  g^n(y)  \tilde{\varphi}(x-y),
    \end{aligned}
\end{equation}
where we defined
\begin{equation}\label{eq:predictor-general-2d}
    \begin{aligned}
    g^n(x) = \sum_{j=1}^n (2\pi g_j) \delta(y-x_j).
    \end{aligned}
\end{equation}
Both for feature regime and NTK limit, the first derivative of $\tilde{\varphi}(x)$ is continuous except for two values of $x$ (0 and $\pi$ for lazy, $-\pi/2$ and $\pi/2$ for feature), so that $\tilde{\varphi}(x)''$ has a singular part consisting of two Dirac delta functions.

As a result, the second derivative of the predictor $(f^n)''$ has a singular part consisting of many Dirac deltas. If we denote with $(f^n)''_r$ the regular part, obtained by subtracting all the delta functions, we can show that (see~\autoref{app:2drig}):

\begin{proposition}\label{prop:derivative} \textbf{(informal)}
As ${n\to \infty}$, $(f^n)_r''$ converges to a function having finite second moment, i.e.
\begin{equation}
    \lim_{n \to \infty}  \mathbb{E}_{f^*}[(f^n)_r''(x)]^2 = \text{const.} < \infty.
\end{equation}
\end{proposition}
 In the large $n$ limit, the predictor displays a singular second derivative at $O(n)$ points. Proposition~\ref{prop:derivative} implies that outside of these singular points the second derivative is well defined.
 Thus, as $n$ gets large and the singular points approach each other, the predictor can be approximated by a chain of parabolas, as highlighted in \autoref{fig:parabolas} and noticed in \cite{tomasini2022failure} for a Laplace kernel. This property alone allows to determine the asymptotic scaling of the error in $d\,{=}\,2$. In simple terms, \autoref{prop:derivative} follows from the convergence of $g^n$ to the function satisfying $f^*(x)\,{=}\,\int\frac{dy}{2\pi} g(y) \tilde{\varphi}_r(x-y)$, which is guaranteed under our assumptions on the target function---a detailed proof is given in~\autoref{app:2drig}.

\paragraph{Decay of the error in $d\,{=}\,2$ (sketch)} The full calculation is in~\autoref{app:2dex}. Consider a slightly simplified problem where $\tilde{\varphi}$ has a single discontinuity in its derivative, located  at $x=0$. In this  case, $f^n(x)$ is singular if and only if $x$ is a data point. Consider then the interval $x\in [x_i, x_{i+1}]$ and set $\delta_i=x_{i+1}-x_i$, $x_{i+1/2}=(x_{i+1}+x_i)/2$. If the target function is smooth enough ($\nu_t\,{>}\,2$), then a Taylor expansion implies $|f^*(x_{i+1/2})-f^n(x_{i+1/2})|\sim \delta_i^2$. Since the distances $\delta_i$ between adjacent singular points are random variables with mean of order $1/n$ and finite moments, it is straightforward to obtain that $\overline{\epsilon}(n)\sim \sum_i (f^*(x_{i+1/2})-f^n(x_{i+1/2}))^2\sim \sum_i \delta_i^4 \sim  n^{-4}$. By contrast if $f^*$ is not sufficiently smooth ($\nu_t\,{\leq}\,2$), then  $|f^*(x_{i+1/2})-f^n(x_{i+1/2})|\sim \delta_i^{2\nu_t}$, leading to $\overline{\epsilon}(n)\sim n^{-2\nu_t}$.  Note that for this  asymptotic argument to apply to the feature learning regime, one must ensure that the distribution of the rescaled distance between adjacent singularities $n \delta_i$ has a finite fourth moment. This is obvious in the lazy regime, where the $\delta_i$'s are controlled by the position of the training points, but not in the feature regime, where the distribution of singular points is determined by that of the neuron's features. Nevertheless, we show that it must be the case in our setup in~\autoref{app:2dex}.

\paragraph{Interpretation in terms of spectral bias}
From the discussion above it is evident that there is a length scale $\delta$ of order $1/n$ such that $f^n(x)$ is a good approximation of $f^*(x)$ over scales larger than $\delta$. In terms of Fourier modes\footnote{The Fourier transform of a function $f(x)$ is indicated by the hat, $\widehat f (k)$.}, one has: \emph{i)} $\widehat{f^n}(k)$ matches $\widehat{f^n}(k)$ at long wavelengths, i.e. for $k\ll k_c \sim 1/n$. \emph{ii}) In addition, since the phases $\exp(i k x_j)$ become effectively random phases for $k\gg k_c$,   $\widehat{g^n}(k)\,{=}\,\sum_j g_j \exp(ikx_j)$ becomes a Gaussian random variable with zero mean and fixed variance and thus \emph{iii)} $\widehat{f^n}(k)\,{=}\,\widehat{g^n}(k)\widehat{\tilde{\varphi}}(k)$ decorrelates from $f^*$ for $k\gg k_c$. Therefore
\begin{equation}
    \overline{\epsilon}(n)\sim \sum_{|k|>k_c}  \mathbb{E}_{f^*}\lSq \lRo \widehat{g^n}(k) \widehat{\tilde{\varphi}}(k)-\widehat{f^n}(k)\rRo^2\rSq \sim  \sum_{|k|\geq k_c} \mathbb{E}_{f^*}\lSq(\widehat{g^n}(k))^2\rSq \widehat{\tilde{\varphi}}(k)^2 +\mathbb{E}_{f^*}\lSq (\widehat{f^n}(k))^2\rSq.
\end{equation} 
 For $\nu_t\,{>}\, 2$, one has  $\sum_j g_j^2\sim n^{-1}\lim_{n\rightarrow \infty}\int g^n(x)^2dx\sim n^{-1}$.  It follows (see ~\autoref{app:spectral} for details) that the sum is dominated by the first term, hence entirely controlled by the  Fourier coefficients of $\widehat{f^n}(k)$ at large $k$. A smoother predictor  corresponds to a faster decay of $\widehat{f^n}(k)$ with $k$, thus a faster decay of the error with $n$. Plugging the relevant decays yields $\overline\epsilon\sim n^{-4}$ for feature regime and lazy regime with the NTK, and $n^{-6}$ for lazy regime with the RFK (which is smoother than the NTK). 
For $\nu_t\,{\leq}\, 2$, the two terms have comparable magnitude (see ~\autoref{app:spectral}), thus $\overline\epsilon\sim n^{-2\nu_t}$.

\paragraph{Generalization to higher dimensions} The argument above can be generalized for any $d$ by replacing Fourier modes with projections onto spherical harmonics. The characteristic distance between training points scales as $n^{-1/(d-1)}$, thus $k_c\sim n^{-1/(d-1)}$. Our ansatz is that, as in $d\,{=}\,2$: \emph{i)} for $k\ll k_c$, the predictor modes coincide with those of the target function, $f^n_{k,l}\approx f^*_{k,l}$ (this corresponds to the spectral bias result of kernel methods, stating that the predictor reproduces the first $O(n)$ projections of the target in the kernel eigenbasis  \cite{bordelon2020spectrum}); \emph{ii)} For $k\gg k_c$, $g^n_{k,l}$ is a sum of uncorrelated terms, thus a Gaussian variable with zero mean and fixed variance; \emph{iii)} $f^n_{k,\ell}\,{=}\,g^n_{k,\ell}\tilde{\varphi}_k$ decorrelates from $f^*_{k,\ell}$ for $k\gg k_c$. \emph{i)}, \emph{ii)} and \emph{iii)} imply that:
\begin{equation} 
   \overline\epsilon(n) \sim \sum_{k\geq k_c} \sum_{l=1}^{\Nc_{k,d}} \mathbb{E}_{f^*}\lSq\lRo f^n_{k,l}-f^*_{k,l}\rRo^2\rSq \sim \sum_{k\geq k_c} \sum_{l=1}^{\Nc_{k,d}} \mathbb{E}_{f^*}\lSq (g^n_{k,l})^2\rSq \varphi_k^2+ k^{-2\nu_t-(d-1)}.
\end{equation}
As shown in~\autoref{app:spectral}, from this expression it is straightforward to obtain~\autoref{mainr}. Notice again that when the target is sufficiently smooth so that the predictor-dependent term dominates, the error is determined by the smoothness of the predictor. In particular, as $d\,{>}\,2$, the predictor of feature learning is less smooth than both the NTK and RFK ones, due to the slower decay of the corresponding $\varphi_k$.

\section{Numerical tests of the theory}\label{sec:numerical}

\begin{figure}[h]
    \centering
    \includegraphics[width=\linewidth]{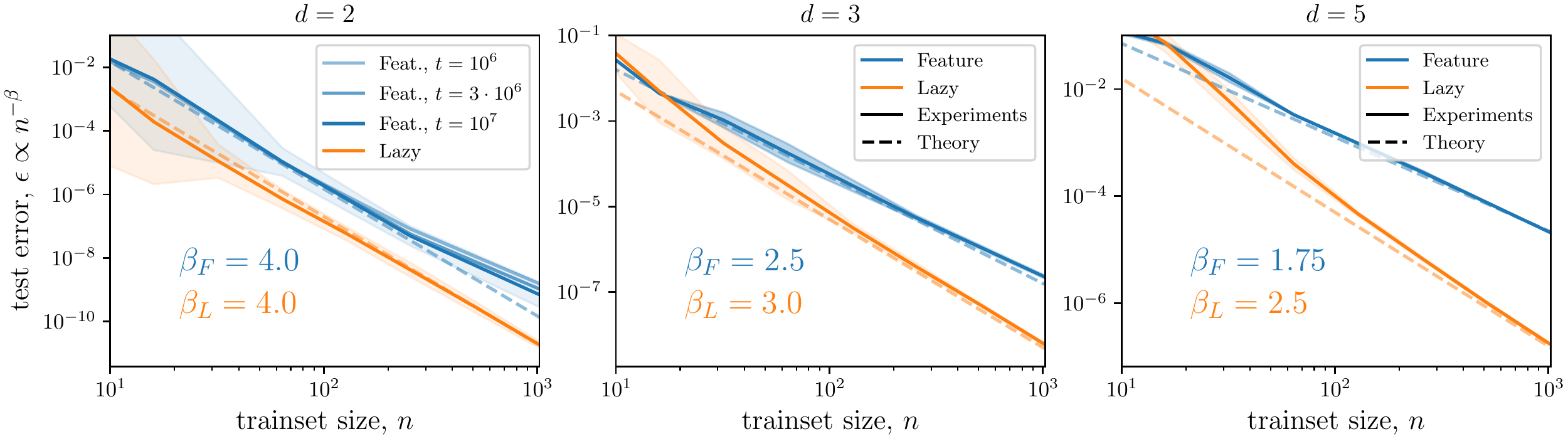}
    \caption{\textbf{Generalization error for a constant function $f^*(\bm{x})=1$.} Generalization error as a function of the training set size $n$ for a network trained in the feature regime with L1 regularization (blue) and kernel regression corresponding to the infinite-width lazy regime (orange). Numerical results (full lines) and the exponents predicted by the theory (dashed) are plotted. 
    Panels correspond to different input-space dimensions ($d = 2, 3, 5$). Results are averaged over 10 different initializations of the networks and datasets. For $d=2$ and large $n$, the gap between experiments and prediction for the feature regime is due to the finite training time $t$. Indeed our predictions become more accurate as $t$ increases, as illustrated in the left. }
    \label{fig:learning_curves}
\end{figure}

\begin{figure}[ht]
\begin{subfigure}{.75\textwidth}
  \centering
\includegraphics[width=1.04\linewidth]{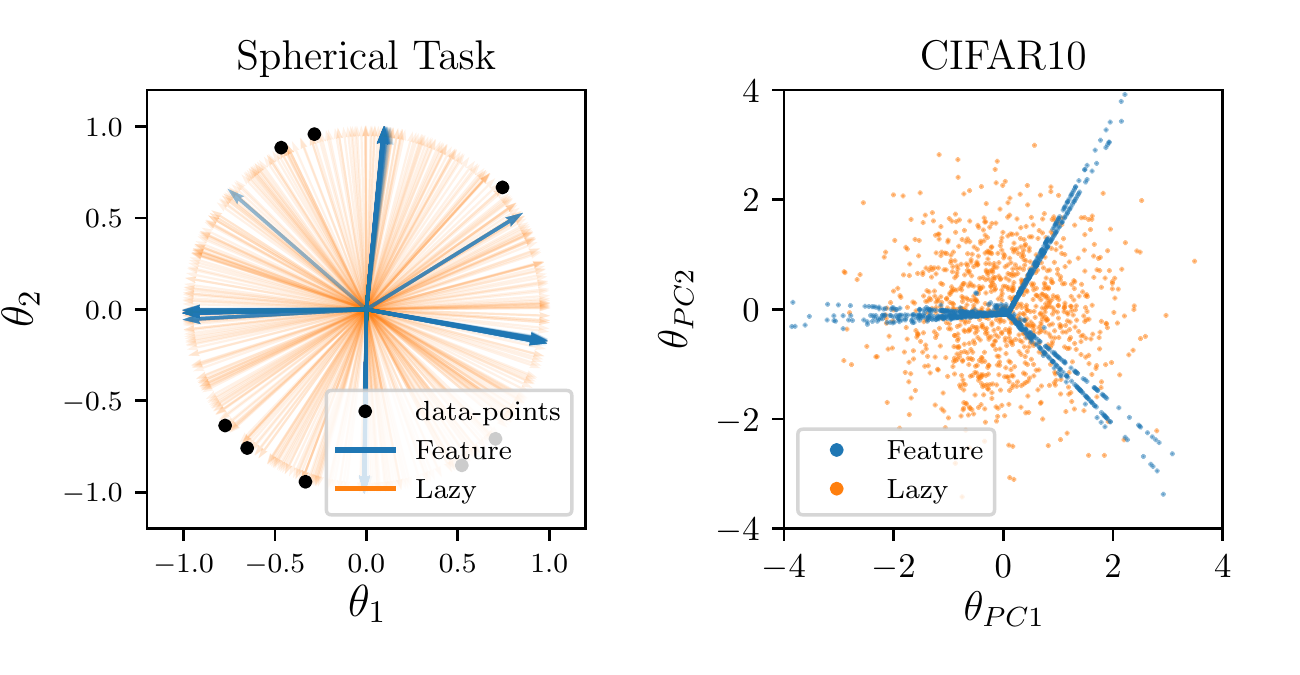} 
  \caption{\textbf{Features sparsification.} \fst Panel: Distribution of neuron's feature for the task of learning a constant function on the sphere in 2D. Arrows represent a subset of the network features $\{\bm{\theta}_h\}_{h=1}^H$ after training in the lazy and feature regimes. Training is performed on $n = 8$ data-points (black dots). \snd Panel: FCN trained on CIFAR10. On the axes the first two principal components of the features $\{\bm{\theta}_h\}_{h=1}^H$ after training on $n=32$ points in the feature (blue) and lazy (orange) regimes. Similarly to what is observed when learning a constant function, the $\bm \theta_h$ angular distribution becomes sparse with training in the feature regime.}
  \label{fig:features_sparsification_tr}
\end{subfigure}\hfill
\begin{subfigure}{.221\textwidth}
  \centering
  \vspace{.38cm}
  \includegraphics[width=.8\linewidth]{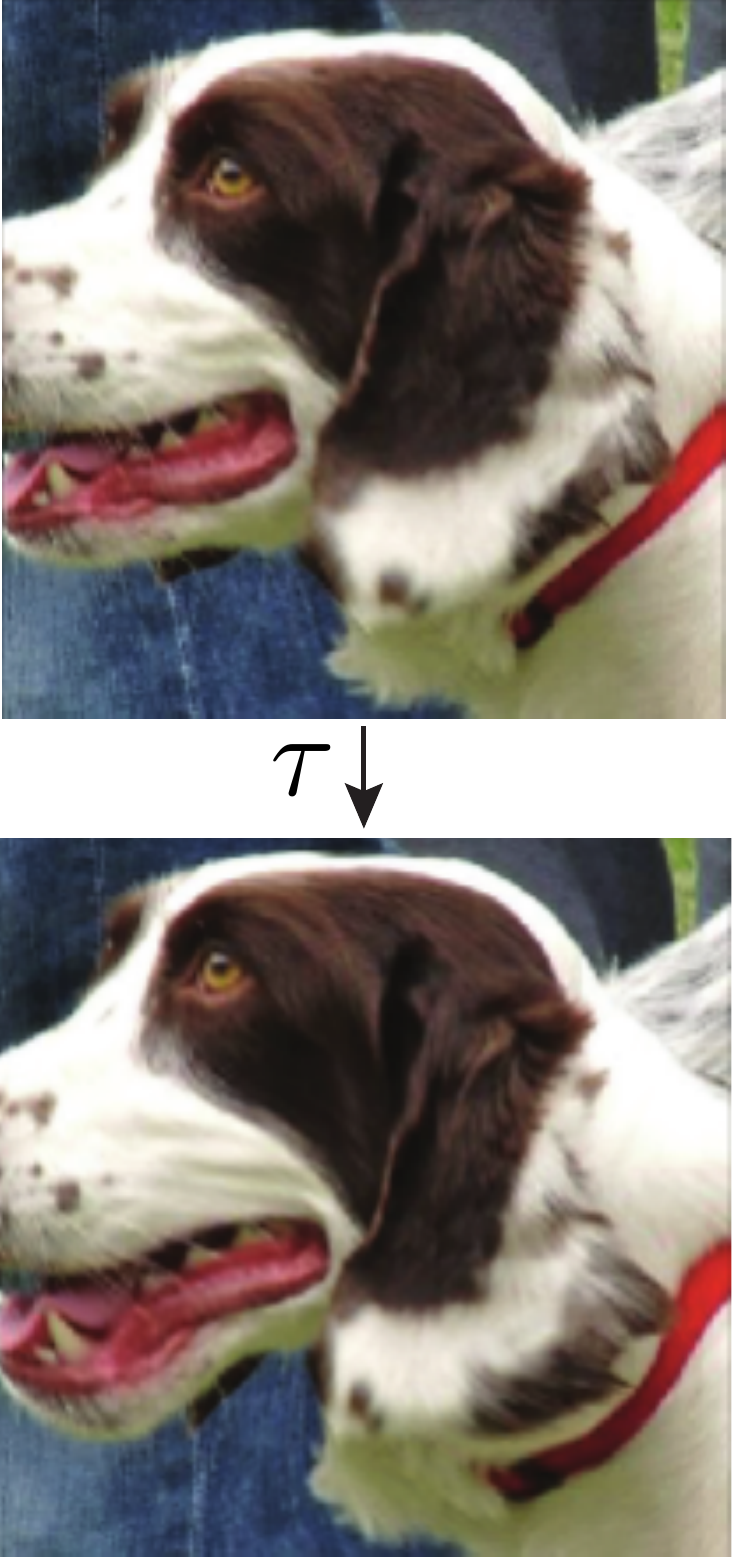}  
  \caption{\textbf{Example of  diffeomorphism.} Sample of a max-entropy deformation $\tau$ \cite{petrini_relative_2021} when applied to a natural image, illustrating that it does not change the image class for the human brain.}
  \label{fig:diffeo_dog}
\end{subfigure}
\caption{\textbf{Features sparsification and example of a diffeomorphism.}}
\label{fig:features_and_diffeo}
\end{figure}

We test successfully our predictions by computing the learning curves of both lazy and feature regimes when \emph{(i)} the target function is constant on the sphere for varying $d$, see \autoref{fig:learning_curves}, and \emph{(ii)} the target is a Gaussian random field with varying smoothness $\nu_t$, as shown in \autoref{fig:grf_lc} of \autoref{app:minnorm_and_gd}. For the lazy regime, we perform kernel regression using the analytical expression of the NTK~\cite{Cho2009} (see also~\autoref{eq:ntk}).
For the feature regime, we find that our predictions hold when having a small regularization, although it takes unreachable times for gradient descent to exactly recover the minimal-norm solution---a more in-depth discussion can be found in \autoref{app:minnorm_and_gd}. An example of the atomic distribution of neurons found after training, which contrasts with the initial distribution, is displayed in \autoref{fig:features_sparsification_tr}, left panel.

Another way to obtain sparse features is to initialize the network with very small weights \cite{woodworth2020kernel}, as proposed in \cite{chizat2019lazy}.  As in the presence of an infinitesimal weights decay, this scheme also leads to sparse solutions with $n_A = \mathcal{O}(n)$ -- an asymptotic dependence confirmed in \autoref{fig:alpha-trick} of \autoref{app:minnorm_and_gd}. This observation implies that our predictions must apply in that case too, as we confirm in \autoref{fig:alpha-trick}.

\section{Evidence for overfitting along diffeomorphisms in image datasets}
\label{sec:evidence_images}

For fully-connected networks, the feature regime is well-adapted to learn anisotropic tasks \cite{chizat2020implicit}: if the target function does not depend on a certain linear subspace of input space, e.g. the pixels at the corner of an image, then neurons align perpendicularly to these directions~\cite{paccolat2020compressing}.
By contrast, our results highlight a drawback of this regime when the target function is constant or smooth along  directions in input space that require a continuous distribution of neurons to be represented. In such a case, the adaptation of the weights to the training points leads to a predictor with a sparse representation. Such a predictor would be less smooth than in the lazy regime and thus underperform.

Does this view hold for images, and explain why learning their features is detrimental for fully-connected networks? The first positive empirical evidence is that the neurons' distribution of networks trained on image data becomes indeed sparse in the feature regime, as illustrated in \autoref{fig:features_sparsification_tr}, right, for CIFAR10 \cite{krizhevsky_learning_2009}. This observation raises the question of which are the directions in input space \mbox{\emph{i)} along} which the target should vary smoothly, and \emph{ii)} that are not easily represented by a discrete set of neurons. An example of such directions are global translations, which conserve the norm of the input and do not change the image class: the lazy regime predictor is indeed smoother than the feature one with respect to translations of the input (see~\autoref{app:sensitivity_other_transf}). Yet, these transformations live in a space of dimension $2$, which is small in comparison with the full dimensionality $d$ of the data and thus may play a negligible role.

A much larger class of transformations believed to have little effect on the target are small diffeomorphisms~\cite{bruna2013invariant}. A diffeomorphism $\tau$ acting on an image is illustrated in \autoref{fig:diffeo_dog}, which highlights that our brain still perceives the content of the transformed image as in the original one. Near-invariance of the task to these transformations is believed to play a key role in the success of deep learning, and in explaining how neural networks beat the curse of dimensionality \cite{mallat2016understanding}. Indeed, if modern architectures can become insensitive to these transformations, then the dimensionality of the problem is considerably reduced. In fact, it was found that the architectures displaying the best performance are precisely those which learn to vary smoothly along such transformations   \cite{petrini_relative_2021}.

Small diffeomorphisms are likely the directions we are looking for. To test this hypothesis, following~\cite{petrini_relative_2021}, we characterize the smoothness of a function along such diffeomorphisms, relative to that of random directions in input space. Specifically, we use the \textit{relative sensitivity}:
\begin{equation}
    R_{f} = \frac{\mathbb E_{x, \tau} \|f(\tau x) - f(x)\|^2}{\mathbb{E}_{x, \eta} \|f(x + \eta) - f(x)\|^2}.
    \label{eq:R_f}
\end{equation}
In the numerator, the average is made over the test set and over an ensemble of diffeomorphisms, reviewed in \autoref{app:diffeo}. The magnitude of the diffeomorphisms is chosen so that each pixel is shifted by one on average. In the denominator, the average runs over the test set and the vectors $\eta$ sampled uniformly on the sphere of radius $\|\eta\|=\mathbb E_{x, \tau} \|\tau x - x\|$, and this fixes the transformations magnitude.

\begin{figure}
    \centering
    \includegraphics[width=\linewidth]{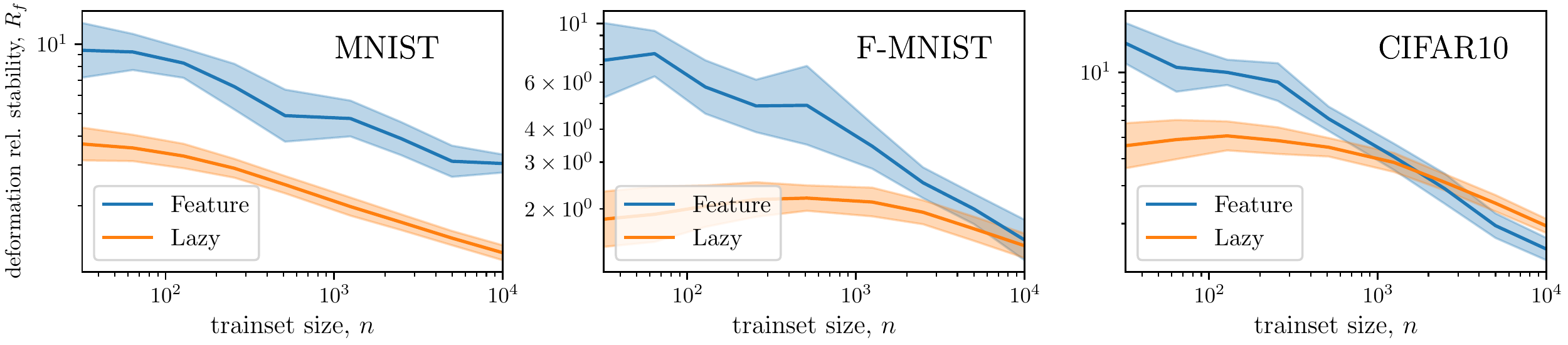}
    \caption{\textbf{Sensitivity to diffeomorphisms vs number of training points.} 
    Relative sensitivity of the predictor to small diffeomorphisms of the input images, in the two regimes, for varying number of training points $n$ and different image datasets. Smaller values correspond to a smoother predictor, on average. Results are computed using the same predictors as in \autoref{fig:learning_curves_images}.}
    \label{fig:stability_vs_n}
\end{figure}

We measure $R_{f}$ as a function of $n$ for three benchmark datasets of images, as shown in \autoref{fig:stability_vs_n}.
We indeed find that $R_{f}$ is consistently smaller in the lazy training regime, where features are not learned.  Overall, this observation supports the view that learning sparse features is detrimental when data present (near) invariance to transformations that cannot be represented sparsely by the architecture considered. \autoref{fig:learning_curves_images} supports the idea that---for benchmark image datasets---this negative effect overcomes well-known positive effects of learning features, e.g. becoming insensitive to pixels on the edge of images (see \autoref{app:sensitivity_other_transf} for evidence of this effect).

\section{Conclusion} 

Our central result is that learning sparse features can be detrimental if the task presents invariance or smooth variations along transformations that are not adequately captured by the neural network architecture. For fully-connected networks, these transformations can be rotations of the input, but also continuous translations and diffeomorphisms. 

Our analysis relies on the sparsity of the features learned by a shallow fully-connected architecture: even in the infinite width limit, when trained in the feature learning regime such networks behave as ${\cal O}(n)$ neurons. The asymptotic analysis we perform for random Gaussian fields on the sphere leads to  predictions for the learning curve exponent $\beta$ in different training regimes, which we verify. Such kind of results is scarce in the literature. 

Note that our analysis focuses on ReLU neurons because \emph{(i)} these are very often used in practice and \emph{(ii)} in that case, $\beta$ will depend on the training regime, allowing for stringent numerical tests. If smooth activations (e.g. softplus) are considered, we expect that learning features will still be detrimental for generalization. Yet, the difference  will not appear in the exponent $\beta$, but in other aspects of the learning curves (including numerical coefficients and pre-asymptotic effects) that are harder to predict.

Most fundamentally, our results underline that the success of feature learning for modern architectures still lacks a sufficient explanation. Indeed, most of the theoretical studies that previously emphasized the benefits of learning features have been considering fully-connected networks, for which learning features can be in practice a drawback. It is tempting to argue that in modern architectures, learning features is not at a disadvantage because smoothness along diffeomorphisms can be enforced from the start---due to the locally connected, convolutional, and pooling layers~\cite{bietti2019group,bruna2013invariant}. Yet the best architectures often do not perform pooling and are not stable toward diffeomorphisms  at initialization. {\it During training}, learning features leads to more stable and smoother solutions along diffeomorphisms  \cite{ruderman_pooling_2018,petrini_relative_2021}. Understanding why building sparse features enhances stability in these architectures may ultimately explain the magical feat of deep CNNs: learning tasks in high dimensions.

\section*{Acknowledgements}
We thank Lénaïc Chizat, Antonio Sclocchi, and Umberto M. Tomasini for helpful discussions. The work of MW is supported by a grant from the Simons Foundation (\#454953) and from the
NSF under Grant No. 200021-165509. The work of EVE is supported by the National Science Foundation  under awards DMR-1420073, DMS-2012510, and DMS-2134216, by the Simons Collaboration on Wave Turbulence, Grant No. 617006, and by a Vannevar Bush Faculty Fellowship.

\bibliography{main}
\bibliographystyle{unsrt}
% \bibliographystyle{apalike}

% \newpage 
% \input{checklist}
\newpage

\input{appendix}

\end{document}

%% file: appendix.tex
\appendix
\counterwithin{figure}{section}

\section{Quick recap of spherical harmonics}\label{app:spherical}

\paragraph{Spherical harmonics} This appendix collects some introductory background on spherical harmonics and dot-product kernels on the sphere~\cite{smola2000regularization}. See~\cite{atkinson2012spherical, efthimiou2014spherical} for an expanded treatment. Spherical harmonics are homogeneous polynomials on the sphere $\mathbb{S}^{d-1}\,{=}\,\lbrace \bm{x}\in\mathbb{R}^d\,|\,\lVert \bm{x} \rVert\,{=}\,1\rbrace$, with $\lVert .\rVert$ denoting the L2 norm. Given the polynomial degree $k\in\mathbb{N}$, there are $\mathcal{N}_{k,s}$ linearly independent spherical harmonics of degree $k$ on $\mathbb{S}^{s-1}$, with
\begin{equation}\label{eq:s-h-mult}
\mathcal{N}_{k,d} = \frac{2k+d-2}{k}\binom{d+k-3}{k-1},\quad \left\lbrace\begin{aligned} &\mathcal{N}_{0,d}=1\quad\forall d,\\ &\mathcal{N}_{k,d}\asymp A_d k^{d-2}\quad\text{for } k\gg 1, \end{aligned}\right.
\end{equation}
where $\asymp$ means logarithmic equivalence for $k\to\infty$ and $A_d = \sqrt{2/\pi} (d-2)^{\frac32-d} e^{d-2}$.
Thus, we can introduce a set of $\mathcal{N}_{k,d}$ spherical harmonics $Y_{k,\ell}$ for each $k$, with $\ell$ ranging in $1,\dots,\mathcal{N}_{k,d}$, which are orthonormal with respect to the uniform measure on the sphere $d\tau(\bm{x})$,
\begin{equation}\label{eq:spherical-harmonics}
    \left\lbrace Y_{k,\ell} \right\rbrace_{k\geq 0,\ell=1,\dots,\Nc_{k,d}},\quad
    \left\langle Y_{k,\ell}, Y_{k,\ell'} \right\rangle_{\sphere{d}} := \int_{\mathbb{S}^{d-1}} Y_{k,\ell}(\bm{x}) Y_{k,\ell'}(\bm{x})\,d\tau(\bm{x}) = \delta_{\ell,\ell'}.
\end{equation}
Because of the orthogonality of homogeneous polynomials with different degree, the set is a complete orthonormal basis for the space of square-integrable functions on $\sphere{d}$. For any function \mbox{$f:\sphere{d}\to\mathbb{R}$}, then
\begin{equation}
    f(\bm{x}) = \sum_{k\geq 0}\sum_{\ell=1}^{\Nc_{k,d}} f_{k,\ell}Y_{k,\ell}(\bm{x}), \quad f_{k,\ell} = \int_{\sphere{d}} f(\bm{x})Y_{k,\ell}(\bm{x})d\tau(\bm{x}).
\end{equation}
Furthermore, spherical harmonics are eigenfunctions of the Laplace-Beltrami operator $\Delta$, which is nothing but the restriction of the standard Laplace operator to $\mathbb{S}^{d-1}$,
\begin{equation}\label{eq:laplac-beltrami}
    \Delta Y_{k,\ell} = -k(k+d-2)Y_{k,\ell}.
\end{equation}

\paragraph{Legendre polynomials} By fixing a direction $\bm{y}$ in $\mathbb{S}^{d-1}$ one can select, for each $k$, the only spherical harmonic of degree $k$ which is invariant for rotations that leave $\bm{y}$ unchanged. This particular spherical harmonic is, in fact, a function of $\bm{x}\cdot\bm{y}$ and is called the Legendre polynomial of degree $k$, $P_{k,d}(\bm{x}\cdot\bm{y})$ (also referred to as Gegenbauer polynomial). Legendre polynomials can be written as a combination of the orthonormal spherical harmonics $Y_{k,\ell}$ via the addition theorem~\cite[Thm.~2.9]{atkinson2012spherical},
\begin{equation}\label{eq:addition}
    P_{k,d}(\bm{x}\cdot\bm{y}) = \frac{1}{\mathcal{N}_{k,d}} \sum_{\ell=1}^{\mathcal{N}_{k,d}} Y_{k,\ell}(\bm{x})Y_{k,\ell}(\bm{y}).
\end{equation}
Alternatively, $P_{k,d}$ is given explicitly as a function of $t\,{=}\,\bm{x}\cdot\bm{y}\in[-1,1]$ via the Rodrigues' formula~\cite[Thm.~2.23]{atkinson2012spherical},
\begin{equation}\label{eq:rodrigues}
    P_{k,d}(t) = \left(-\frac{1}{2}\right)^k \frac{\Gamma\left(\frac{d-1}{2}\right)}{\Gamma\left(k+\frac{d-1}{2}\right)}\left(1-t^2\right)^{\frac{3-d}{2}} \frac{d^k}{dt^k}\left(1-t^2\right)^{k+\frac{d-3}{2}}.
\end{equation}
Here $\Gamma$ denotes the Gamma function, $\Gamma(z)\,{=}\,\int_0^\infty x^{z-1}e^{-x}\,dx$.
Legendre polynomials are orthogonal on $[-1,1]$ with respect to the measure with density $(1-t^2)^{(d-3)/2}$, which is the probability density function of the scalar product between to points on $\sphere{d}$.
\begin{equation}\label{eq:legendre-ortho}
    \int_{-1}^{+1} P_{k,d}(t)P_{k',d}(t)\,\left(1-t^2\right)^{\frac{d-3}{2}}dt = \frac{|\mathbb{S}^{d-1}|}{|\mathbb{S}^{d-2}|}\frac{\delta_{k,k'}}{\mathcal{N}_{k,s}}.
\end{equation}
Here $|\mathbb{S}^{d-1}|\,{=}\,2\pi^{\frac d2}/\Gamma(\frac d2)$ denotes the surface area of the $d$-dimensional unit sphere ($|\mathbb{S}^{0}|\,{=}\,2$ by definition).

To sum up, given $\bm{x},\bm{y}\in\mathbb{S}^{d-1}$, functions of $\bm{x}$ or $\bm{y}$ can be expressed as a sum of projections on the orthonormal spherical harmonics, whereas functions of $\bm{x}\cdot\bm{y}$ can be expressed as a sum of projections on the Legendre polynomials. The relationship between the two expansions is elucidated in the Funk-Hecke formula~\cite[Thm.~2.22]{atkinson2012spherical},
\begin{equation}\label{eq:funk-hecke}
    \int_{\mathbb{S}^{d-1}} f(\bm{x}\cdot\bm{y}) Y_{k,\ell}(\bm{y})\,d\tau(\bm{y}) = Y_{k,\ell}(\bm{x})\frac{|\mathbb{S}^{d-2}|}{|\mathbb{S}^{d-1}|}\int_{-1}^{+1}  f(t) P_{k,d}(t)\,\left(1-t^2\right)^{\frac{d-3}{2}}dt:= f_k Y_{k,\ell}(\bm{x}).
\end{equation}

\subsection{Expansion of ReLU and combinations thereof}\label{aapp:relu-harmonic}

We can apply~\autoref{eq:funk-hecke} to have an expansion of neurons $\sigma\lRo \bm{\theta}\cdot\bm{x}\rRo$ in terms of spherical harmonics~\cite[Appendix~D]{bach2017breaking}. After defining
\begin{equation}
    \varphi_k := \frac{|\mathbb{S}^{d-2}|}{|\mathbb{S}^{d-1}|}\int_{-1}^{+1} \sigma(t) P_{k,d}(t)\,\left(1-t^2\right)^{\frac{d-3}{2}}dt,
\end{equation}
one has
\begin{equation}\label{eq:relu-mercer}
 \sigma\lRo \bm{\theta}\cdot\bm{x}\rRo = \sum_{k\geq 0} \Nc_{k,d}\varphi_k P_{k,d}\lRo \bm{\theta}\cdot\bm{x}\rRo = \sum_{k\geq 0} \varphi_k \sum_{\ell=1}^{\Nc_{k,d}} Y_{k,\ell}(\bm{\theta})Y_{k,\ell}(\bm{x}).
\end{equation}
For ReLU activations, in particular, $\sigma(t)\,{=}\,\text{max}(0,t)$, thus
\begin{equation}\label{eq:relu-eigval}
    \varphi_k^{\text{ReLU}} = \frac{|\mathbb{S}^{d-2}|}{|\mathbb{S}^{d-1}|}\int_{0}^{+1} t P_{k,d}(t)\,\left(1-t^2\right)^{\frac{d-3}{2}}dt.
\end{equation}
Notice that when $k$ is odd $P_{k,d}$ is an odd function of $t$, thus the integrand $t P_{k,d}(t)(1-t^2)^{\frac{d-3}{2}}$ is an even function of $t$. As a result the integral on the right-hand side of~\autoref{eq:relu-eigval} coincides with half the integral over the full domain $[-1,1]$,
\begin{equation}
    \int_{0}^{+1} t P_{k,d}(t)\,\left(1-t^2\right)^{\frac{d-3}{2}}dt = \frac{1}{2}\int_{-1}^{+1} t P_{k,d}(t)\,\left(1-t^2\right)^{\frac{d-3}{2}}dt = 0\,\text{ for } k>1, 
\end{equation}
because, due to~\autoref{eq:legendre-ortho}, $P_{k,d}$ is orthogonal to all polynomials with degree strictly lower than $k$. For even $k$ we can use~\autoref{eq:rodrigues} and get~\cite{bach2017breaking}
(see~\autoref{eq:unit-decay}, main text)
\begin{equation}\begin{aligned}
     \int_{0}^{+1} t P_{k,d}(t)\,\left(1-t^2\right)^{\frac{d-3}{2}}dt &= \left(-\frac{1}{2}\right)^k \frac{\Gamma\left(\frac{d-1}{2}\right)}{\Gamma\left(k+\frac{d-1}{2}\right)}\int_0^1 t \frac{d^k}{dt^k}\left(1-t^2\right)^{k+\frac{d-3}{2}} \,dt\\
    &=-\left(-\frac{1}{2}\right)^k \frac{\Gamma\left(\frac{d-1}{2}\right)}{\Gamma\left(k+\frac{d-1}{2}\right)}\left. \frac{d^{k-2}}{dt^{k-2}}\left(1-t^2\right)^{k+\frac{d-3}{2}} \right\rvert_{t=0}^{t=1}\\
    &\Rightarrow \varphi_k^{\text{ReLU}} \sim k^{-\frac{d-1}{2}-\frac{3}{2}}\,\text{ for }\,k\gg 1\text{ and even.}
\end{aligned}\end{equation}
Because all $\varphi^{\text{ReLU}}_k$ with $k\,{>}\,1$ and odd vanish, even summing an infinite amount of neurons $\sigma(\bm{\theta}\cdot\bm{x})$ with varying $\bm{\theta}$ does not allow to approximate any function on $\sphere{d}$, but only those which have vanishing projections on all the spherical harmonics $Y_{k,\ell}$ with $k\,{>}\,1$ and odd. This is why we set the odd coefficients of the target function spectrum to zero in~\autoref{eq:target}.

\subsection{Dot-product kernels on the sphere}
Also general dot-product kernels on the sphere admit an expansion such as~\autoref{eq:relu-mercer},
\begin{equation}\label{eq:kernel-mercer}
 \mathcal{C}\lRo \bm{x}\cdot\bm{y}\rRo = \sum_{k\geq 0} \Nc_{k,d}c_k P_{k,d}\lRo \bm{\theta}\cdot\bm{x}\rRo = \sum_{k\geq 0} c_k \sum_{\ell=1}^{\Nc_{k,d}} Y_{k,\ell}(\bm{\theta})Y_{k,\ell}(\bm{x}),
\end{equation}
with
\begin{equation}\label{eq:ker-eigvals}
c_k =\frac{|\mathbb{S}^{d-2}|}{|\mathbb{S}^{d}|}\int_{-1}^1 \mathcal{C}(t)P_{k,d}(t)\,\left(1-t^2\right)^{\frac{d-3}{2}}dt.
\end{equation}
The asymptotic decay of $c_k$ for large $k$ is controlled by the behaviour of $\mathcal{C}(t)$ near $t\,{=}\,\pm 1,$~\cite{bietti_deep_2021}. More precisely~\cite[Thm.~1]{bietti_deep_2021}, if $\mathcal{C}$ is infinitely differentiable in $(-1,1)$ and has the following expansion around $\pm 1$,
\begin{equation}\label{eq:biettibach}
\left\lbrace\begin{aligned}
    \mathcal{C}(t) &= p_1(1-t) + c_1(1-t)^\nu + o\lRo(1-t)^\nu\rRo\text{ near }t=+1;\\
    \mathcal{C}(t) &= p_{-1}(-1+t) + c_{-1}(-1+t)^\nu + o\lRo(-1+t)^\nu\rRo\text{ near }t=-1,
\end{aligned}\right.\end{equation}
where $p_{\pm1}$ are polynomials and $\nu$ is not an integer, then
\begin{equation}\label{eq:eigvals-decay}
\begin{aligned}
    k\text{ even: }& c_k \sim (c_1 + c_{-1}) k^{-2\nu-(d-1)};\\
    k\text{ odd: }& c_k \sim (c_1 - c_{-1}) k^{-2\nu-(d-1)},
\end{aligned}\end{equation}
The result above implies that that if $c_1\,{=}\,c_{-1}$ ($c_1\,{=}\,-c_{-1}$), then the eigenvalues with $k$ odd (even) decay faster than $k^{-2\nu-(d-2)}$. Moreover, if $\mathcal{C}$ is infinitely differentiable in $[-1,1]$ then $c_k$ decays faster than any polynomial.

\paragraph{NTK and RFK of one-hidden-layer ReLU networks}
Let $\mathbb{E}_{\bm{\theta}}$ denote expectation over a multivariate normal distribution with zero mean and unitary covariance matrix. For any $\bm{x}$, $\bm{y}\in\sphere{d}$, the RFK of a one-hidden-layer ReLU network~\autoref{eq:model} with all parameters initialised as independent Gaussian random numbers with zero mean and unit variance reads
\begin{equation}\label{eq:rfk}
\begin{aligned}
    K^{\text{RFK}}(\bm{x}\cdot\bm{y}) &= \mathbb{E}_{\bm{\theta}}\lSq \sigma(\bm{\theta}\cdot\bm{x})\sigma(\bm{\theta}\cdot\bm{y}) \rSq\\
    &= \frac{(\pi-\arccos{(t)})t + \sqrt{1-t^2}}{2\pi},\text{ with }\,t=\bm{x}\cdot\bm{y}.
\end{aligned}
\end{equation}
The NTK of the same network reads, with $\sigma'$ denoting the derivative of ReLU or Heaviside function,
\begin{equation}\label{eq:ntk}
\begin{aligned}
    K^{\text{NTK}}(\bm{x}\cdot\bm{y}) &= \mathbb{E}_{\bm{\theta}}\lSq \sigma(\bm{\theta}\cdot\bm{x})\sigma(\bm{\theta}\cdot\bm{y}) \rSq + (\bm{x}\cdot\bm{y})\mathbb{E}_{\bm{\theta}}\lSq \sigma'(\bm{\theta}\cdot\bm{x})\sigma'(\bm{\theta}\cdot\bm{y}) \rSq\\
    &= \frac{2(\pi-\arccos{(t)})t + \sqrt{1-t^2}}{2\pi},\text{ with }\,t=\bm{x}\cdot\bm{y}.
\end{aligned}
\end{equation}
As functions of a dot-product on the sphere, both NTK and RFK admit a decomposition in terms of spherical harmonics as~\autoref{eq:ker-eigvals}. For dot-product kernels, this expansion coincides with the Mercer's decomposition of the kernel~\cite{smola2000regularization}, that is the coefficients of the expansion are the eigenvalues of the kernel. The asymptotic decay of the eigenvalues of such kernels $\varphi_k^{\text{NTK}}$ and $\varphi_k^{\text{RFK}}$ can be obtained by applying~\autoref{eq:biettibach} \cite[Thm.~1]{bietti_deep_2021}. Equivalently, one can notice that $K^{\text{RFK}}$ is proportional to the convolution on the sphere of ReLU with itself, therefore $\varphi_k^{\text{RFK}}\,{=}\,(\varphi_k^{\text{ReLU}})^2$. Similarly, the asymptotic decay of $\varphi_k^{\text{NTK}}$ can be related to that of the coefficients of $\sigma'$, derivative of ReLU: $\varphi_k(\sigma') \sim k \varphi(\sigma)$, thus $\varphi_k^{\text{NTK}}\sim k^2(\varphi_k^{\text{ReLU}})^2$.  Both methods lead to~\autoref{eq:unit-decay} of the main text.

\paragraph{Gaussian random fields and~\autoref{eq:diff-via-kernel}} Consider a Gaussian random field $f^*$ on the sphere with covariance kernel $\mathcal{C}(\bm{x}\cdot\bm{y})$,
\begin{equation}\label{eq:def-grf}
    \expec{f^*(\bm{x})}=0,\quad\expec{f^*(\bm{x})f^*(\bm{y})}=\mathcal{C}(\bm{x}\cdot\bm{y}),\quad \forall\bm{x},\bm{y}\in\sphere{d}.
\end{equation}
$f^*$ can be equivalently specified via the statistics of the coefficients $f^*_{k,\ell}$,
\begin{equation}
    \expec{f^*_{k,\ell}}=0,\quad \expec{f^*_{k,\ell}f^*_{k',\ell'}} = c_k\delta_{k,k'}\delta_{\ell,\ell'},
\end{equation}
with $c_k$ denoting the eigenvalues of $\mathcal{C}$ in~\autoref{eq:ker-eigvals}. Notice that the eigenvalues are degenerate with respect to $\ell$ because the covariance kernel is a function $\bm{x}\cdot\bm{y}$: as a result, the random function $f^*$ is isotropic in law.

If $c_k$ decays as a power of $k$, then such power controls the weak differentiability (in the mean-squared sense) of the random field $f^*$. In fact, from~\autoref{eq:laplac-beltrami},
\begin{equation}
    \norm{\Delta ^{m/2} f^*} = \sum_{k\geq 0}\sum_{\ell} \left(-k(k+d-2)\right)^m \left(f^*_{k,\ell}\right)^2.
\end{equation}
Upon averaging over $f^*$ one gets
\begin{equation}\label{eq:msd1}
\mathbb{E}\lSq\norm{\Delta ^{m/2} f^*}\rSq = \sum_{k\geq 0} \left(-k(k+d-2)\right)^m \sum_{\ell}\mathbb{E}\lSq\left(f^*_{k,\ell}\right)^2\rSq = \sum_{k\geq 0} \left(-k(k+d-2)\right)^m \mathcal{N}_{k,d} c_k.
\end{equation}
From~\autoref{eq:biettibach} \cite[Thm.~1]{bietti_deep_2021}, if $\mathcal{C}(t) \sim (1-t)^{\nu_t}$ for $t\to 1$ and/or $\mathcal{C}(t) \sim (-1+t)^{\nu_t}$ for $t\to -1$, then $c_k\sim k^{-2\nu_t -(d-1)}$ for $k\gg 1$. In addition, for finite but arbitrary $d$, $\left(-k(k+d-2)\right)^m\sim k^{2m}$ and $\mathcal{N}_{k,s}\sim k^{d-2}$ (see~\autoref{eq:s-h-mult}). Hence the summand in the right-hand side of~\autoref{eq:msd1} is $\sim k^{2(m-\nu_t)-1}$, thus
\begin{equation}
    \mathbb{E}\lSq\norm{\Delta ^{m/2} f^*}\rSq < \infty\quad\forall m<\nu_t.
\end{equation}
Alternatively, one can think of $\nu_t$ as controlling the scaling of the difference $\delta f^*$ over inputs separated by a distance $\delta$. From~\autoref{eq:def-grf},
\begin{equation}\begin{aligned}
    \expec{\lvert f^*(\bm{x})-f^*(\bm{y}) \rvert^2} &= 2\mathcal{C}(1)-2\mathcal{C}(\bm{x}\cdot\bm{y}) = 2\mathcal{C}(1) + O((1-\bm{x}\cdot\bm{y})^{\nu_t})\\
    &= 2\mathcal{C}(1) + O(\lvert\bm{x}-\bm{y}\rvert^{2\nu_t})
\end{aligned}\end{equation}

\section{Uniqueness and Sparsity of the L1 minimizer}\label{app:l1minimization}

Recall that we want to find the $\gamma^*$ that solves
\begin{equation}
       \gamma^*= \argmin_{\gamma} \int_{\sphere{d}} \lvert d\gamma(\bm{\theta})\rvert \quad \text{subject to} \quad  \int_{\sphere{d}} \,\sigma(\bm{\theta}\cdot\bm{x}_i)d\gamma(\bm{\theta}){=}f^*(\bm{x}_i)\quad \forall i=1,\dots,n.
       \label{eq:predictor-feature-app}
\end{equation}
In this appendix, we argue that the uniqueness of $\gamma^*$ which implies that it is atomic with at most $n$ atoms is a natural assumption. We start by discretizing the measure $\gamma$ into $H$ atoms, with $H$ arbitrarily large. Then the problem \autoref{eq:predictor-feature-app} can be rewritten as 
\begin{equation}
    \bm w^* = \argmin_{\bm w} \|\bm w\|_1, \quad \text{subject to} \quad \bm\Phi \bm w = \bm y,
    \label{eq:l1-min-problem}
\end{equation}
with $\bm \Phi \in \mathbb{R}^{H\times n}$, $\Phi_{h, i} = \sigma(\bm \theta_h \cdot \bm x_i)$ and $y_i = f^*(\bm x_i)$. 

Given $\bm w \in \mathbb{R}^H$, let $\bm u = \max(\bm w, 0)\ge \bm 0$ and $\bm v = - \max (-\bm w,0)\ge \bm 0$ so that $\bm w = \bm u - \bm v$. It is well-known (see e.g. ~\cite{chen1998atomic})
that the minimization problem in~\eqref{eq:l1-min-problem} can be recast in terms of $\bm u$ and $\bm v$ into a linear programming problem. That is, $\bm w^* = \bm u^* - \bm v^*$ with
\begin{equation}
    (\bm u^*,\bm v^*) = \argmin_{\bm u, \bm v} \bm e^T (\bm u + \bm v),\quad \text{subject to} \ \bm\Phi \bm u - \bm\Phi \bm v= \bm y, \quad \bm u \ge \bm 0, \quad \bm v \ge \bm 0
    \label{eq:lp-problem}
\end{equation}
where $\bm e = [1,1, \ldots,1]^T$.
Assuming that this problem is feasible (i.e. there is at least one solution to $\bm\Phi \bm u - \bm\Phi \bm v= \bm y$ such that $\bm u \ge \bm 0$, $\bm v \ge \bm 0$), it is known that it admits extremal solution, i.e. solutions such that at most $n$ entries of $(\bm u^*,\bm v^*)$ (and hence $\bm w^*$) are non-zero. The issue is whether such an extremal solution is unique. Assume that there are two, say $(\bm u^*_1, \bm v^*_1)$ and $(\bm u^*_2, \bm v^*_2)$. Then, by convexity,  
\begin{equation}
    (\bm u^*_t, \bm v^*_t ) = (\bm u^*_1, \bm v^*_1 ) t + (\bm u^*_2, \bm v^*_2) (1-t) 
\end{equation}
is also a minimizer of \eqref{eq:lp-problem}  for all $t\in [0,1]$, with the same minimum value $\bm u^*_t+ \bm v^*_t= \bm u^*_1+ \bm v^*_1= \bm u^*_2+ \bm v^*_2$. Generalizing this argument to the case of more than two extremal solutions, we conclude that all minimizers are global, with the same minimum value, and they live on the simplex where $\bm e^T (\bm u + \bm v) = \bm e^T (\bm u_1 + \bm v_1)$. Therefore, nonuniqueness requires that that this simplex has a nontrivial intersection with the feasible set where $\bm\Phi \bm u - \bm\Phi \bm v= \bm y$ with $\bm u \ge \bm 0$, $\bm v \ge \bm 0$. We argue that, generically, this will not be the case, i.e. the intersection will be trivial, and the extremal solution unique. In particular, since in our case  we are in fact interested in the problem~\eqref{eq:predictor-feature-app},  we can always perturb slightly the discretization into $H$ atoms of $\gamma$ to guarantee that the extremal solution is unique. Since this is true no matter how large $H$ is, and any Radon measure can be approached to arbitrary precision using such discretization, we conclude that the minimizer of~\eqref{eq:predictor-feature-app} should be unique as well, with at most $n$ atoms.

\section{Proof of Proposition~\ref{prop:derivative}}\label{app:2drig}

In this section, we provide the formal statement and proof of Proposition~\ref{prop:derivative}. Let us recall the general form of the predictor for both lazy and feature regimes in $d\,{=}\,2$. From~\autoref{eq:predictor-general-2d},
\begin{equation}\label{eq:proof1-predictor}
    f^n(x) = \sum_{j=1}^n g_j \tilde{\varphi}(x-x_j) =\int \frac{dy}{2\pi}  g^n(y)  \tilde{\varphi}(x-y).
\end{equation}
where $n$ is the number of training points for the lazy regime and the number of atoms for the feature regime and, for $x\in(-\pi,\pi]$,
\begin{equation}\label{eq:-proof1-predictor}
    \tilde{\varphi}(x) = \left\lbrace\begin{aligned}
    \text{max}\lCu 0,\cos{(x)}\rCu\quad&\text{(feature regime)},\\ 
    \frac{2(\pi-\lvert x \rvert)\cos(x)+\sin(\lvert x\rvert)}{2\pi}\quad&\text{(lazy regime, NTK)},\\
    \frac{(\pi-\lvert x \rvert)\cos(x)+\sin(\lvert x\rvert)}{2\pi}\quad&\text{(lazy regime, RFK)}.
\end{aligned}\right.
\end{equation}
All these functions $\tilde{\varphi}$ have jump discontinuities on some derivative: the first for feature and NTK, the third for RFK. If the $l$-th derivative has jump discontinuities, the $l\,{+}\,1$-th only exists in a distributional sense and it can be generically written as a sum of a regular function and a sequence of Dirac masses located at the discontinuities. With $m$ denoting the number of such discontinuities and $\lbrace x_j\rbrace_j$ their locations, $f^{(l)}$ denoting the $l$-th derivative of $f$, for some $c_j\in\mathbb{R}$,
\begin{equation}\label{eq:proof1-singular}
    f^{(l+1)}(x) = f^{(l+1)}_r(x) + \sum_{j=1}^m c_j \delta(x-x_j),
\end{equation}
where $f_r$ denotes the \emph{regular} part of $f$.

\begin{customprop}{2} Consider a random target function $f^*$ satisfying~\autoref{eq:target} and the predictor $f^n$ obtained by training a one-hidden-layer ReLU network on $n$ samples $(x_i,f^*(x_i))$ in the feature or in the lazy regime~(\autoref{eq:proof1-predictor}). 
Then, with $\widehat f(k)$ denoting the Fourier transform of $f(x)$, one has
\begin{equation}
% \begin{align}
\lim_{|k|\to\infty} \lim_{n\to \infty} \frac{\widehat{(f^n)''_r}(k)}{\widehat{f^*}(k)} = c,\label{eq:proof1-featntk}%\\ \text{\it (RFK): }&\lim_{k\to\infty} \lim_{n\to \infty} \frac{(f^n)^{(4)}_r(k)}{f^*(k)} = c,\label{eq:proof1-rfk}
% \end{align}
\end{equation}
where $c$ is a constant (different for every regime). This result implies that as ${n\to \infty}$, $(f^n)''(x)$ converges to  a function having finite second moment, i.e.
\begin{equation}\begin{aligned}
    \lim_{n \to \infty}  \mathbb{E}_{f^*}\lSq(f^n)_r''(x)\rSq^2 &= \lim_{n \to \infty}  \mathbb{E}_{f^*}\lSq\int dx \lRo(f^n)_r''\rRo^2(x)\rSq \\&= \lim_{n \to \infty}  \mathbb{E}_{f^*}\lSq\sum_k\widehat{(f^n)''_r}^2(k) \rSq = \text{const.} < \infty,
\end{aligned}\end{equation}
 using the fact that $\mathbb{E}_{f^*}[(f^n)_r''(x)]^2$ does not depend on $x$ and $\mathbb{E}_{f^*}[\sum_k\widehat{(f^*)}^2(k)]=\text{const}$.
\end{customprop}

{\it Proof:} Because our target functions  are random fields that are in $L_2$ with probability one, and the RKHS of our kernels are dense in that space, we know that the test error vanishes as $n\rightarrow \infty$ \cite{bach_learning_2022}. As a result
\begin{equation}\label{eq:proof1-limit}
   f^*(x) = \lim_{n\to\infty} f^n(x) = \lim_{n\to\infty} \int \frac{dy}{2\pi}  g^n(y)  \tilde{\varphi}(x-y).
\end{equation}
Consider first the feature regime and the NTK lazy regime. In both cases $\tilde{\varphi}$ has two jump discontinuities in the first derivative, located at $x\,{=}\,0,\pi$ for the NTK and at $x\,{=}\,\pm \pi/2$, therefore we can write the second derivative as the sum of a regular function and two Dirac masses,
\begin{equation}\begin{aligned}
    (\tilde{\varphi}^{\text{FEATURE}})'' &=  -\text{max}\lCu 0,\cos{(x)}\rCu+ \delta(x-\pi/2) + \delta(x+\pi/2),\\
    (\tilde{\varphi}^{\text{NTK}})'' &=  \frac{-2(\pi-\lvert x\rvert)\cos(x)+3\sin(\lvert x\rvert)}{2\pi}- \frac{1}{2\pi}\delta(x) + \frac{1}{2\pi}\delta(x-\pi).
\end{aligned}\end{equation}
As a result, the second derivative of the predictor can be written as the sum of a regular part $(f^n)''_r$ and a sequence of $2n$ Dirac masses. After subtracting the Dirac masses, both sides of~\autoref{eq:proof1-predictor} can be differentiated twice and yield
\begin{equation}\label{eq:proof1-deriv-real}
    (f^n)''_r(x) = \int \frac{dy}{2\pi}  g^n(y)  \tilde{\varphi}_r''(x-y).
\end{equation}
Hence in the Fourier representation we have
\begin{equation}
    \label{eq:s1}
    \widehat{(f^n)''_r}(k) = \widehat{g^n}(k) (-k^2 \widehat{\tilde{\varphi}}_r(k))
\end{equation}
where we defined
\begin{equation}
    \widehat{\tilde{\varphi}}(k) = \int_{-\pi}^{\pi}\frac{d x }{\sqrt{2\pi}}e^{ikx}\tilde{\varphi} (x), \qquad \widehat{\tilde{\varphi}_r}(k) = \int_{-\pi}^{\pi}\frac{d x }{\sqrt{2\pi}}e^{ikx}\tilde{\varphi}_r (x).
\end{equation}
and used $\widehat{\tilde{\varphi}_r''}(k) = -k^2 \widehat{\tilde{\varphi}_r}(k)$.
By universal approximation we have
\begin{equation}
\label{eq:s2}
    \widehat{f^*}(k) = \int_{-\pi}^{\pi}\frac{d x }{\sqrt{2\pi}}e^{ikx}f^* (x)= \lim_{n\to\infty} \widehat{g^n}(k)  \widehat{\tilde{\varphi}}(k) \qquad \Rightarrow \qquad \lim_{n\to\infty} \widehat{g^n}(k) =  \frac{\widehat{f^*}(k)}{\widehat{\tilde{\varphi}}(k)}.
\end{equation}
As a result by combining \autoref{eq:s1} and \autoref{eq:s2} we deduce
\begin{equation}
    \lim_{n\to\infty} \widehat{(f^n)''_r}(k) = -\frac{k^2 \widehat{\tilde{\varphi}}_r(k)}{\widehat{\tilde{\varphi}}(k)} \widehat{f^*}(k).
\end{equation}

To complete the proof using this result it remains to estimate the scaling of $\widehat{\tilde{\varphi}}_r(k)$ and $\widehat{\tilde{\varphi}}(k)$ in the large $|k|$ limit.

For the feature regime, a direct calculation shows that $\tilde{\varphi}''_r\,{=}\,-\tilde{\varphi}$, implying that $\widehat{\tilde{\varphi}}_r(k)=-\widehat{\tilde{\varphi}}(k)$. This proves  that~\autoref{eq:proof1-featntk} is satisfied with $c\,{=}\,-1$.

For the NTK lazy regime $\tilde{\varphi}''_r$ and $-\tilde{\varphi}$ are different but they have similar singular expansions near $x\,{=}\,0$ and $\pi$. Therefore their Fourier coefficients display the same asymptotic decay. More specifically, with $t\,{=}\,\cos(x)$ (or $x\,{=}\,\arccos(t)$), so that $\tilde{\varphi}(x)\,{=}\,\varphi(t)$, one has
\begin{equation}
\left\lbrace\begin{aligned}
    \varphi^{\text{NTK}}(t) &= t - \frac{1}{\sqrt{2}\pi}(1-t)^{1/2} + O\lRo(1-t)^{3/2}\rRo\text{ near }t=+1;\\
    \varphi^{\text{NTK}}(t) &= - \frac{1}{\sqrt{2}\pi}(-1+t)^{1/2} + O\lRo(-1+t)^{3/2}\rRo\text{ near }t=-1,
\end{aligned}\right.\end{equation}
and
\begin{equation}
\left\lbrace\begin{aligned}
    (\varphi^{\text{NTK}})''_r(t) &= -t + \frac{5}{\sqrt{2}\pi}(1-t)^{1/2} + O\lRo(1-t)^{3/2}\rRo\text{ near }t=+1;\\
    (\varphi^{\text{NTK}})''_r(t) &= + \frac{5}{\sqrt{2}\pi}(-1+t)^{1/2} + O\lRo(-1+t)^{3/2}\rRo\text{ near }t=-1.
\end{aligned}\right.\end{equation}
Therefore, due to~\autoref{eq:eigvals-decay}, ~\autoref{eq:proof1-featntk} is satisfied with $c\,{=}\,-5$. The same procedure can be applied to the RFK lazy regime, with the exception that it is the fourth derivative of $\tilde{\varphi}^{\text{RFK}}$ which can be written as a regular part plus Dirac masses, but one can still obtain the Fourier coefficients of the second derivative's regular part by dividing those of the fourth derivative's regular part by $k^2$.

\section{Asymptotics of generalization in $d\,{=}\,2$}\label{app:2dex}

In this section we compute the decay of generalization error $\overline\epsilon$ with the number of samples $n$ in the following $2$-dimensional setting:
\begin{equation}
    f^n(x) = \sum_{j=1}^n g_j \tilde{\varphi}(x-x_j),
\end{equation}
where the $x_j$'s are the training points (like in the NTK case) and $\varphi$ has a single discontinuity on the first derivative in $0$. 

Let us order the training points clockwise on the ring, such that $x_1\,{=}\,0$ and $x_{i+1}>x_{i}$ for all $i\,{=}\,1,\dots,n$, with $x_{n+1}\,{:=}\,2\pi$. On each of the $x_i$ the predictor coincides with the target,
\begin{equation}\label{eq:2d-conditions}
    f^n(x_i)=f^*(x_i)\quad \forall\,i=1,\dots,n.
\end{equation}
For large enough $n$, the difference $x_{i+1}-x_i$ is small enough such that, within $(x_i,x_{i+1})$, $f^n(x)$ can be replaced with its Taylor series expansion up to the second order. In practice, the predictors appear like the cable of a suspension bridge with the pillars located on the training points. In particular, we can consider an expansion around $x_i^+\,{:=}\,x_i+\epsilon$ for any $\epsilon\,{>}\,0$ and then let $\epsilon\to 0$ from above:
\begin{equation}\label{eq:2d-taylor-exp-1}
    f^n(x)=f^n(x_i^+) + (x-x_i^+) {f^n}'(x_i^+) + \frac{(x-x_i^+)^2}{2} (f^n)''(x_i^+) + \mathcal{O}\lRo (x-x_i^+)^3 \rRo.
%    f^n(x)&=f^n(x_{i+1}^-) - (x_{i+1}^--x_i) f^n'(x_{i+1}^-) + \frac{(x_{i+1}^--x_i)^2}{2} f^n''(x_{i+1}^-) + \mathcal{O}\lRo (x_{i+1}^--x_i)^3 \rRo.
\end{equation}
By differentiability of $f^n$ in $(x_i,x_{i+1})$ the second derivative can be computed at any point inside $(x_i,x_{i+1})$ without changing the order of approximation in~\autoref{eq:2d-taylor-exp-1}, in particular we can replace $ (f^n)''(x_i^+)$ with $c_{i}$, the mean curvature of $f^n$ in $(x_i,x_{i+1})$. Moreover, as $\epsilon\to0$, $f^n(x_i^+)\to f^*(x_i)$ and $f^n(x_{i+1}^-)\to f^*(x_{i+1})$. By introducing the limiting slope $m_i^{+}\,{:=}\,\lim_{x\to 0^+} {f^n}'(x_i+ x)$, we can write
\begin{equation}\label{eq:2d-taylor-exp-2}
    f^n(x)=f^*(x_i) + (x-x_i) m_i^+ + \frac{(x-x_i)^2}{2} c_{i} + O\lRo (x-x_i^+)^3 \rRo%\\    f^n(x)&=f^*(x_{i+1}) - (x_{i+1}-x) m_{i+1}^- + \frac{(x_{i+1}-x)^2}{2} c_{i,i+1} + \mathcal{O}\lRo (x_{i+1}^--x)^3 \rRo.\label{eq:2d-taylor-b}
\end{equation}
Computing~\autoref{eq:2d-taylor-exp-2} at $x\,{=}\,x_{i+1}$ yields a closed form for the limiting slope $m_i^{+}$ as a function of the mean curvature $c_{i}$, the interval length $\delta_i\,{:=}\,(x_{i+1}-x_i)$ and $\Delta f_{i}\,{:=}\,f^*(x_{i+1})-f^*(x_{i})$.  %Coupling these with~\autoref{eq:2d-conditions} yields the $q_i$'s up to corrections in $\mathcal{O}\lRo (x_{i+1}-x_i)^2\rRo$.
Specifically,
\begin{equation}
   m_{i}^+ = \frac{\Delta f_i}{\delta_i} - \frac{\delta_i}{2} c_i.
\end{equation}

The generalization error can then be split into contributions from all the intervals.  If $\nu_t>2$, A Taylor expansion leads to:
\begin{equation}\begin{aligned}
    &\epsilon(n) = \int_0^{2\pi} \frac{dx}{2\pi} \lRo f^n(x)-f^*(x)\rRo^2 \\&= \sum_{i=1}^n \int_{x_i}^{x_{i+1}} \frac{dx}{2\pi}\,\lSq (x-x_i) \lRo m_i^+ - (f^*)'(x_i)\rRo + \frac{(x-x_i)^2}{2} \lRo c_{i} - (f^*)''(x_i)\rRo +o\lRo (x-x_i^+)^2\rRo \rSq^2\\
    &= \sum_{i=1}^n \int_{0}^{\delta_i} \frac{d\delta}{2\pi}\,\lSq \delta \lRo m_i^+ - (f^*)'(x_i)\rRo + \frac{\delta^2}{2} \lRo c_{i} - (f^*)''(x_i)\rRo +o\lRo \delta^2\rRo \rSq^2 \\
    &= \sum_{i=1}^n \frac{1}{2\pi}\left[ \frac{\delta_i^3}{3}\lRo m_i^+ - (f^*)'(x_i)\rRo^2 + \frac{\delta_i^5}{20}\lRo c_{i} - (f^*)''(x_i)\rRo^2\right.\\ &\qquad\qquad\qquad\qquad\left.+\frac{\delta_i^4}{4}\lRo m_i^+ - (f^*)'(x_i)\rRo\lRo c_{i} - (f^*)''(x_i)\rRo +o(\delta_i^5)\right].
\end{aligned}\end{equation}
In addition, as $\Delta f_i \,{=}\, (f^*)'(x_i)\delta_i + (f^*)''(x_i)\delta_i ^2/2 + O(\delta_i^3)$,
\begin{equation}
    m_{i}^+ - (f^*)'(x_i) = \frac{\delta_i}{2}\lRo (f^*)''(x_i)-c_i\rRo + o(\delta_i)^2,
\end{equation}
thus
\begin{equation}
\epsilon(n) = \frac{1}{2\pi} \sum_{i=1}^n \left[\frac{\delta_i^5}{120}\lRo  c_i - (f^*)''(x_i)\rRo^2 + o(\delta_i^5)\right].
\end{equation}
implying:

\begin{equation}
\overline{\epsilon}(n) = \frac{n^{-4}{\lRo n^{-1}\sum_{i=1}^n (n\delta_i)^5 \rRo}}{240\pi} \lim_{n\rightarrow\infty}\int  \mathbb{E}_{f^*}\lSq\lRo(f^n)''(x) - (f^*)''(x)\rRo^2\rSq dx + o(n^{-4})\sim \frac{1}{n^4}
\end{equation}
where we used that  \emph{(i)} the integral converges to some finite value, due to proposition 2. From \autoref{app:2drig}, this integral can be estimated  as $\sum_k  \mathbb{E}_{f^*}\lSq\lRo c f^*(k) - k^2f^*(k)\rRo^2\rSq $, that indeed converges for $\nu_t>2$.
\emph{(ii)}  ${\lRo n^{-1}\sum_{i=1}^n (n\delta_i)^5 \rRo}$ has a {deterministic} limit for large $n$. It is clear for the lazy regime since the distance between adjacent singularities $\delta_i$ follows an exponential distribution of mean $\sim \frac{1}{n}$. We expect this result to be also true for the feature regime in our set-up. Indeed, in the limit $n\rightarrow 
\infty$,  the predictor approaches a parabola between singular points, which generically cannot fit more than three random points. There must thus be a singularity at least every two data-points with a probability approaching unity as $n\rightarrow 
\infty$, which implies that ${\lRo n^{-1}\sum_{i=1}^n (n\delta_i)^5 \rRo}$ converges to a constant for large $n$.

Finally, for $\nu_t<2$, the same decomposition in intervals applies, but a Taylor expansion to second order does not hold. The error is then dominated by the fluctuations of $f^*$ on the scale of the intervals, as indicated in the main text.

\section{Asymptotic of generalization via the spectral bias ansatz}\label{app:spectral}

According to the spectral bias ansatz, the first $n$ modes of the predictor $f^n_{k,\ell}$ coincide with the modes of the target function $f^*_{k,\ell}$. Therefore, the asymptotic scaling of the error with $n$ is entirely controlled by the remaining modes, 
\begin{equation}
    \epsilon(n) \sim  \sum_{k\geq k_c} \sum_{\ell=1}^{\Nc_{k,d}} \lRo f^n_{k,\ell}-f^*_{k,\ell}\rRo^2\,\text{ with }\,  \sum_{k\leq k_c}\Nc_{k,d}  \sim n.
\end{equation}
Since $\Nc_{k,d}\sim k^{d-2}$ for $k\gg 1$, one has that, for large $n$, $k_c\sim n^{\frac{1}{d-1}}$. After averaging the error over target functions we get
\begin{equation}
    \overline\epsilon(n) \sim \sum_{k\geq k_c}\sum_{\ell=1}^{\Nc_{k,d}} \left\lbrace \mathbb{E}_{f^*}\lSq\lRo f^n_{k,\ell}\rRo^2\rSq + \mathbb{E}_{f^*}\lSq\lRo f^*_{k,\ell}\rRo^2\rSq - 2\mathbb{E}_{f^*}\lSq\lRo f^n_{k,\ell}f^*_{k,\ell}\rRo\rSq \right\rbrace.
\end{equation}

Let us recall that, with the predictor having the general form in~\autoref{eq:predictor-general}, then
\begin{equation}
    f^n_{k,\ell}=g^n_{k,\ell}\varphi_k \quad\text{ with }\quad g^n_{k,\ell} = \sum_{j=1}^n g_j Y_{k,\ell}(\bm{y}_j),
\end{equation}
where the $\bm{y}_j$'s denote the training points for the lazy regime and the neuron features for the feature regime. For $k\,{\ll}\,k_c$, where $f^n_{k,\ell}\,{=}\,f^*_{k,\ell}$, $g^n_{k,\ell}\,{=}\, f^*_{k,\ell}/\varphi_k$. For $k\,{\gg}\,k_c$, due to the highly oscillating nature of $Y_{k,\ell}$, the factors $Y_{k,\ell}(\bm{y}_j)$ are essentially decorrelated random numbers with zero mean and finite variance, since the values of $(Y_{k,\ell}(\bm{y}_j))^2$ are limited by the addition theorem~\autoref{eq:addition}. Let us denote the variance with $\sigma_Y$. By the central limit theorem, $g^n_{k,\ell}$ converges to a Gaussian random variable with zero mean and finite variance $\sigma_Y^2 \sum_{j=1}^n g_j^2$. As a result,
\begin{equation}\label{eq:error-spectral}\begin{aligned}
    \overline\epsilon(n) \sim& \sum_{k\geq k_c}\sum_{\ell=1}^{\Nc_{k,d}} \left\lbrace \left(\sum_{j=1}^n g_j^2\right)\varphi_k^2 + \mathbb{E}_{f^*}\lSq\lRo f^*_{k,\ell}\rRo^2\rSq  \right\rbrace\\
    =& \left(\sum_{j=1}^n g_j^2\right) \sum_{k\geq k_c} \Nc_{k,d} \varphi_k^2 + \sum_{k\geq k_c} \Nc_{k,d} c_k,
\end{aligned}\end{equation}
where we have used the definition of $f^*$ (\autoref{eq:target}) to set the expectation of $(f^*_{k,\ell})^2$ to $c_k$. 

\paragraph{Large $\nu_t$ case} When $f^*$ is smooth enough the error is controlled by the predictor term proportional to $\sum_{j=1}^n g_j^2$. More specifically, if
\begin{equation}\label{eq:conv-g}
    \sum_{k\geq 0}\sum_{\ell=1}^{\Nc_{k,d}}  \frac{c_k}{\varphi_k^2}  < +\infty,
\end{equation}
then the function $g^n(\bm{x})$ converges to the square-summable function $g^*(\bm{x})$ such that $f^*(\bm{x})\,{=}\,\int g^*(\bm{y}) \varphi(\bm{x}\cdot\bm{y}) \,d\tau(\bm{y})$. With $c_k\sim k^{-2\nu_t-(d-1)}$ and $\Nc_{k,d}\sim k^{d-2}$, in the lazy regime $\varphi_k\sim k^{-(d-1)-2\nu}$~\autoref{eq:conv-g} is satisfied when $2\nu_t\,{>}\,2(d-1)+4\nu$ ($\nu\,{=}\,1/2$ for the NTK and $3/2$ for the RFK). In the feature regime $\varphi_k\sim k^{-(d-1)/2-3/2}$,~\autoref{eq:conv-g} is satisfied when $2\nu_t\,{>}\,(d-1)+3$. If $g^n(\bm{x})$ converges to a square-summable function, then
\begin{equation}
    \sum_{j=1}^n g_j^2 = \frac{1}{n} \int g^n(\bm{x})^2\,d\tau(\bm{x}) + o(n^{-1}) = \frac{1}{n} \sum_{k\geq 0} \Nc_{k,d}  \frac{c_k}{\varphi_k^2}  + o(n^{-1}) ,
\end{equation}
which is proportional to $n^{-1}$. In addition, since $\Nc_{k,d}\sim k^{d-2}$ and $k_c\sim n^{\frac{1}{d-1}}$, one has
\begin{equation}\label{eq:scaling-smooth}
   n^{-1}\sum_{k\geq k_c} \Nc_{k,d}\varphi_k \sim \left\lbrace\begin{aligned}  &\left.n^{-1}k^{d-1} k^{-2(d-1)-4\nu }\right\rvert_{k=n^{\frac{1}{d-1}}}\sim n^{-2-\frac{4\nu}{d-1}}\text{ (Lazy)},\\&\left.n^{-1}k^{d-1} k^{-(d-1)-3 }\right\rvert_{k=n^{\frac{1}{d-1}}}\sim n^{-1-\frac{3}{d-1}}\text{ (Feature)}, \end{aligned}\right.
\end{equation}
and
\begin{equation}\label{eq:scaling-target}
   \sum_{k\geq k_c} \Nc_{k,d}c_k \left.\sim k^{d-1} k^{-2\nu_t -(d-1)}\right\rvert_{k=n^{\frac{1}{d-1}}} \sim n^{-\frac{2\nu_t}{d-1}}.
\end{equation}
Hence, if $\nu_t$ is large enough so that~\autoref{eq:conv-g} is satisfied, the asymptotic decay of the error is given by~\autoref{eq:scaling-smooth}.

\paragraph{Small $\nu_t$ case} If~\autoref{eq:scaling-smooth} does not hold then $g^n(\bm{x})$ is not square-summable in the limit $n\to\infty$. However, for large but finite $n$ only the modes up to the $k_c$-th are correctly reconstructed, therefore
\begin{equation}\label{eq:scaling-nonsmooth}
    \sum_{j=1}^n g_j^2 \sim  \frac{1}{n} \sum_{k\leq k_c} \Nc_{k,d}  \frac{c_k}{\varphi_k^2} \sim\left\lbrace\begin{aligned}  &\left.n^{-1}k^{-2\nu_t} k^{2(d-1)+4\nu }\right\rvert_{k=n^{\frac{1}{d-1}}}\sim n^{-\frac{2\nu_t}{d-1}}n^{1+\frac{4\nu}{d-1}}\text{ (Lazy)},\\&\left.n^{-1}k^{-2\nu_t} k^{(d-1)+3 }\right\rvert_{k=n^{\frac{1}{d-1}}}\sim n^{-\frac{2\nu_t}{d-1}} n^{\frac{3}{d-1}}\text{ (Feature)}, \end{aligned}\right.
\end{equation}
Both for feature and lazy, multiplying the term above by $\sum_{k\geq k_c} \Nc_{k,d}\varphi_k$ from~\autoref{eq:scaling-smooth} yields $\sim n^{-2\nu_t/(d-1)}$. This is also the scaling of the target function term~\autoref{eq:scaling-target}, implying that for small $\nu_t$ one has
\begin{equation}
   \overline\epsilon(n) \sim n^{-\frac{2\nu_t}{d-1}}
\end{equation}
both in the feature and in the lazy regimes.

\section{Spectral bias via the replica calculation}\label{app:replica}

Due to the equivalence with kernel methods, the asymptotic decay of the test error in the lazy regime can be computed with the formalism of \cite{bordelon2020spectrum}, which also provides a non-rigorous justification for the spectral bias ansatz. By ranking the eigenvalues from the biggest to the smallest, such that $\varphi_{\rho}$ denotes the $\rho$-th eigenvalue and denoting with $c_\rho$ the variance of the projections of the target onto the $\rho$-th eigenfunction, one has
\begin{equation}\label{eq:replica}
    \epsilon(n) =\sum_{\rho} \epsilon_{\rho}(n),\quad \epsilon_\rho(n) = \frac{\kappa(n)^2}{(\varphi_\rho+\kappa(n))^2 }c_\rho,\quad \kappa(n) = \frac{1}{n}\sum_\rho \frac{\varphi_\rho \kappa(n)}{\varphi_\rho + \kappa(n)}.
\end{equation}
It is convenient to introduce the eigenvalue density,
\begin{equation}
    \mathcal{D}(\varphi) := \sum_{k\geq 0}\sum_{l=1}^{\Nc_{k,d}} \delta(\varphi-\varphi_k) = \sum_{k\geq 0} \Nc_{k,d} \delta(\varphi-\varphi_k) \sim \int_{0}^\infty k^{d-2} \delta(\varphi - k^{-(d-1)-2\nu}) \text{ for } k\gg 1.
\end{equation}
After changing variables in the delta function, one finds 
\begin{equation}
    \mathcal{D}(\varphi) \sim \varphi^{-\frac{2(d-1)+2\nu}{(d-1)+2\nu}}\text{ for } \varphi \ll 1.
\end{equation}
This can be used for inferring the asymptotics of $\kappa(n)$,
\begin{equation}\begin{aligned}
    \kappa(n) &= \frac{1}{n}\sum_\rho \frac{\varphi_\rho \kappa(n)}{\varphi_\rho + \kappa(n)} \sim \frac{1}{n}\int d\varphi\,\mathcal{D}(\varphi) \frac{\varphi \kappa(n)}{\varphi + \kappa(n)} \\ &\sim \frac{1}{n}\int_0^{\kappa(n)} d\varphi\, \mathcal{D}(\varphi) \varphi + \frac{\kappa(n)}{n}\int_{\kappa(n)}^{\varphi_0} d\varphi\,\mathcal{D}(\varphi)\\
    &\sim \frac{1}{n}\kappa(n)^{1-\frac{(d-1)}{(d-1)+2\nu}} \Rightarrow \kappa(n) \sim n^{-1-\frac{2\nu}{d-1}}.
\end{aligned}\end{equation}
Once the scaling of $\kappa(n)$ has been determined, the modal contributions to the error can be split according to whether $\varphi_\rho \ll \kappa(n)$ or $\varphi_\rho \gg \kappa(n)$. The scaling of $\varphi_\rho$ with the rank $\rho$ is determined self-consistently,
\begin{equation}
   \rho \sim \int_{\varphi_\rho}^{\varphi_1} d\varphi\,\mathcal{D}(\varphi) \sim \varphi_\rho^{-\frac{d-1}{(d-1)+2\nu}} \Rightarrow \varphi_\rho \sim \rho^{-1-\frac{2\nu}{d-1}} \Rightarrow   \varphi_\rho \gg (\ll) \kappa(n) \Leftrightarrow \rho \ll (\gg) n.
\end{equation}
Therefore
\begin{equation}\label{eq:error-replica}\begin{aligned}
 \epsilon(n) &\sim \kappa(n)^2 \sum_{\rho \ll n} \frac{c_\rho}{\varphi^2_\rho} + \sum_{\rho \gg n} c_\rho.
\end{aligned}\end{equation}
Notice that $\kappa(n)^2$ scales as $n^{-1}\sum_{k\geq k_c} \Nc_{k,s}\varphi_k$ in~\autoref{eq:scaling-smooth}, whereas $\sum_{\rho \ll n} c_\rho/\varphi_\rho^2$ corresponds to $n \sum_{j} g_j^2$ in~\autoref{eq:scaling-nonsmooth}, so that the first term on the right-hand side of~\autoref{eq:error-replica} matches that of~\autoref{eq:error-spectral}. The same matching is found for the second term on the right-hand side of~\autoref{eq:error-replica}, so that the replica calculation justifies the spectral bias ansatz.

\section{Training wide neural networks: does gradient descent (GD) find the minimal-norm solution?}
\label{app:minnorm_and_gd}

In the main text we provided predictions for the asymptotics of the test error of the minimal norm solution that fits all the training data.
Does the prediction hold when solution of \autoref{eq:predictor-feature} and \autoref{eq:predictor-lazy-ntk} is approximately found by GD?
More specifically, is the solution found by GD the minimal-norm one?

\paragraph{Feature Learning}

We answer these questions by performing full-batch gradient descent in two settings (further details about the trainings are provided in the code repository, \texttt{experiments.md} file),

\begin{enumerate}
    \item \textbf{Min-L1.} Here we update weights and features of \autoref{eq:model}, with $\xi = 0$, by following the negative gradient of 
    \begin{equation}
        \mathcal{L}_\text{Min-L1} = \frac{1}{2n}\sum_{i=1}^n\lRo f^*(\bm{x}_i)-f(\bm{x}_i) \rRo^2 + \frac{\lambda}{H}\sum_{h=1}^H |w_h|,
    \end{equation}
    with $\lambda \to 0^+$. The weights $w_h$ are initialized to zero and the features are initialized uniformly and constrained to be on the unit sphere.
    
    \item \textbf{$\alpha$-trick.} Following \cite{chizat2019lazy}, here we minimize
    \begin{equation}
        \mathcal{L}_{\alpha\text{-trick}} = \frac{1}{2n\alpha}\sum_{i=1}^n\lRo f^*(\bm{x}_i)-\alpha f(\bm{x}_i) \rRo^2,
    \end{equation}
    with $\alpha\to 0$. This trick allows to be far from the lazy regime by forcing the weights to evolve to $\mathcal{O}(1/\alpha)$, when fitting a target of order 1.
\end{enumerate}

\begin{figure}
    \centering
    \includegraphics[width=\textwidth]{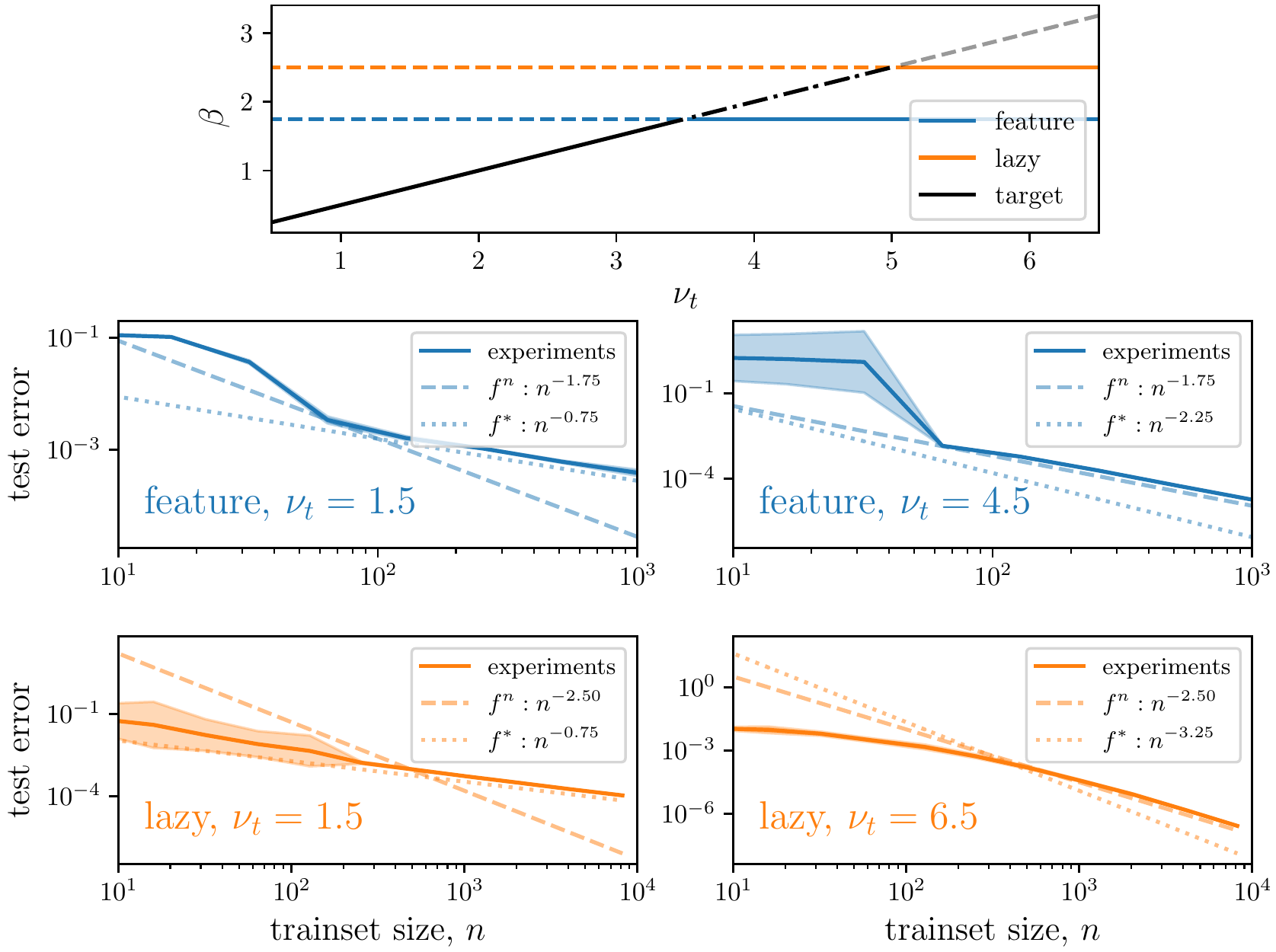}
    \caption{\textbf{Gen. error decay vs. target smoothness and training regime.} Here, data-points are sampled uniformly from the spherical surface in $d=5$ and the target function is an infinite-width FCN with activation function $\sigma(\cdot) = |\cdot|^{\nu_t - \nicefrac{1}{2}}$, corresponding to a Gaussian random process of smoothness $\nu_t$. \fst row: gen. error decay exponent as a function of the target smoothness $\nu_t$. The three curves correspond to the target contribution to the generalization error (black) and the predictor contribution in either feature (blue) or lazy (orange) regime. Full lines highlight the dominating contributions to the gen. error. \snd row: agreement between predictions and experiments in the feature regime for a non-smooth (left) and smooth (right) target. In the first case, the error is dominated by the target $f^*$, in the second by the predictor $f^n$ -- predicted exponents $\beta$ are indicated in the legends. \trd row: analogous of the previous row for the lazy regime.}
    \label{fig:grf_lc}
\end{figure}

\begin{figure}
    \centering
    \includegraphics[width=\linewidth]{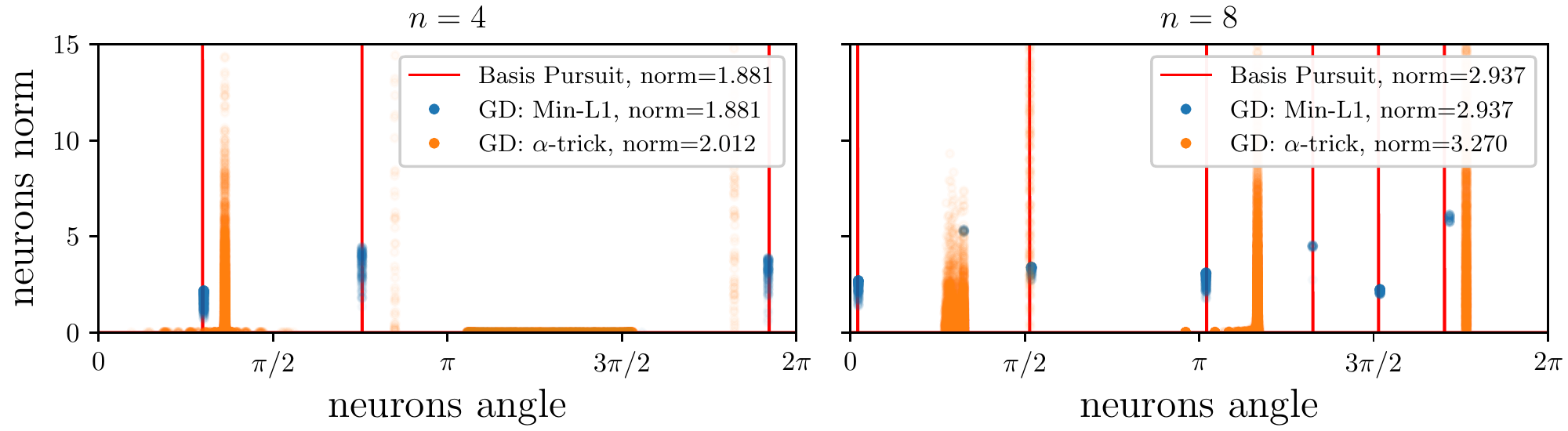}
    \caption{\textbf{Comparing solutions.} Solutions to the spherically symmetric task in $d=2$ for $n=4$ (left) and $n=8$ (right) training points. In red the minimal norm solution (\autoref{eq:predictor-feature}) as found by Basis Pursuit \cite{chen1998atomic}. Solutions found by GD in the Min-L1 and $\alpha$-trick setting are respectively shown in blue and orange. Dots correspond to single neurons in the network. The $x$-axis reports their angular position while the $y$-axis reports their norm: $|w_h|\|\bm \theta_h\|_2$. The total norm of the solutions, $\frac{\alpha}{H}\sum_{h=1}^H|w_h|\|\bm \theta_h\|_2$, is indicated in the legend.}
    \label{fig:minL1sol}
\end{figure}

\begin{figure}
    \centering
    \includegraphics[width=\textwidth]{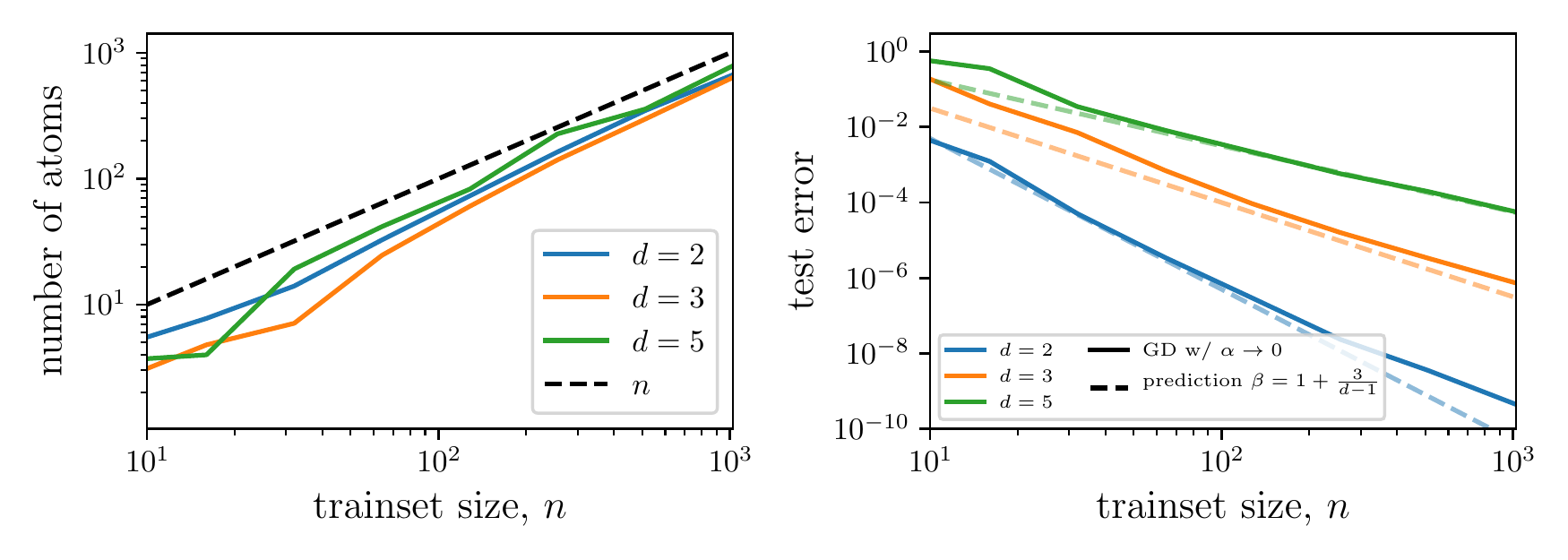}
    \caption{\textbf{Solution found by the $\alpha$-trick.} We consider here the case of approximating the constant target function on $\mathbb S^{d-1}$ with an FCN. Training is performed starting from small initialization through the $\alpha$-trick. Left: Number of atoms $n_A$ as a function of the number of training points $n$. Neurons that are active on the same subset of the training set are grouped together and we consider each group a distinct atom for the counting. Right: Generalization error in the same setting (full), together with the theoretical predictions (dashed). Different colors correspond to different input dimensions. 
    The case of $d=2$ and large $n$ suffers from the same finite time effects discussed in \autoref{fig:learning_curves}.
    Results are averaged over 10 different initializations of the networks and datasets.}
    \label{fig:alpha-trick}
\end{figure}

In both cases, the solution found by GD is sparse, in the sense that is supported on a finite number of neurons -- in other words, the measure $\gamma(\bm \theta)$ becomes atomic, satisfying Assumption \ref{as:sparse:AR}.
Furthermore, we find that
\begin{enumerate}
    \item For \textbf{Min-L1}, the generalization error prediction holds (\autoref{fig:learning_curves} and \autoref{fig:grf_lc}) as the the minimal norm solution if effectively recovered, see \autoref{fig:minL1sol}. Such clean results in terms of features position are difficult to achieve for large $n$ because the training dynamics becomes very slow and reaching convergence becomes computationally infeasible. Still, we observe the test error to plateau and reach its infinite-time limit much earlier than the parameters, which allows for the scaling predictions to hold.
    \item \textbf{$\alpha$-trick}, however, does not recover the minimal-norm solution, \autoref{fig:minL1sol}. Still, the solution found is of the type (\ref{eq:N:atoms}) as it is sparse and supported on a number of atoms that scales linearly with $n$, \autoref{fig:alpha-trick}, left. For this reason, we find that our predictions for the generalization error hold also in this case, see \autoref{fig:alpha-trick}, right.
\end{enumerate}

\paragraph{Lazy Learning}
In this case, the correspondence between the solution found by gradient descent and the minimal-norm one is well established \cite{jacot2018neural}. Therefore, numerical experiments are performed here via kernel regression and the analytical NTK \autoref{eq:ntk}: given a dataset $\{\bm x_i, y_i = f^*(\bm x_i)\}_{i=1}^n$, we define the gram matrix $\mathbf{K} \in \mathbb{R}^{n \times n}$ with elements $\mathbf{K}_{ij} = K(\bm x_i, \bm x_j)$ and the vector of target labels $\bm{y} = [y_1, y_2, \dots, y_n]$. The $q_i$'s in \autoref{eq:sol-lazy} can be easily recovered by solving the linear system 
\begin{equation}
    \bm y = \tfrac{1}{n} \mathbf{K} \bm q .
\end{equation}

\paragraph{Experiments} Numerical experiments are run with PyTorch on GPUs NVIDIA V100 (university internal cluster). Details for reproducing experiments are provided in the
\href{https://github.com/pcsl-epfl/regressionsphere}{code repository}, \href{https://github.com/pcsl-epfl/regressionsphere/blob/main/experiments.md}{\texttt{experiments.md} file}. 
Individual trainings are run in 1 minute to 1 hour of wall time. We estimate a total of a thousand hours of computing time for running the preliminary and actual experiments present in this work.

\newpage
\section{Sensitivity of the predictor to transformations other than diffeomorphisms}
\label{app:sensitivity_other_transf}

This section reports experiments to integrate the discussion of \autoref{sec:evidence_images}. In particular, we: \textit{(i)} show that the lazy regime predictor is less sensitive to image translations than the feature regime one (as is the case for deformations, from~\autoref{fig:stability_vs_n}); \textit{(ii)} provide evidence of the positive effects of learning features in image classifications, namely becoming invariant to pixels at the border of images which are unrelated to the task. 

\begin{figure}
    \centering
    \includegraphics[width=\textwidth]{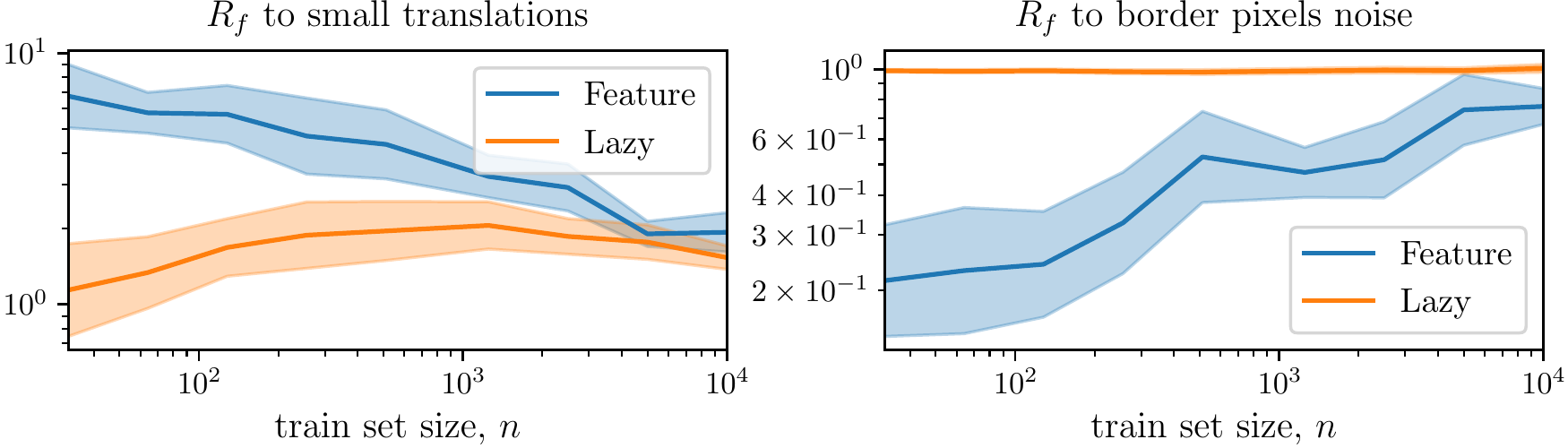}
    \caption{\textbf{Sensitivity to input transformations vs number of training points.} 
    Relative sensitivity of the predictor to (left) random 1-pixel translations and (right) white noise added at the boundary of the input images, in the two regimes, for varying number of training points $n$ and when training on FashionMNIST. Smaller values correspond to a smoother predictor, on average. Results are computed using the same predictors as in \autoref{fig:learning_curves_images}. Left: For small translations, the behavior is the same compared to applying diffeomorphisms. Right: The lazy regime does not distinguish between noise added at the boundary or on the whole image ($R_f = 1$), while the feature regime gets more insensitive to the former.}
    \label{fig:tr_noise_stability}
\end{figure}

To prove the above points we consider, as in~\autoref{fig:stability_vs_n}, the relative sensitivity of the predictors of lazy and feature regime with respect to global translations for point \textit{(i)} and corruption of the boundary pixels for point \textit{(ii)}. The relative sensitivity to translations is obtained from~\autoref{eq:R_f} after replacing the transformation $\tau$ with a one-pixel translation of the image in a random direction. For the relative sensitivity to boundary corruption, the transformation consists in adding zero-mean and unit-variance Gaussian numbers to the boundary pixels. Both relative sensitivities are plotted in \autoref{fig:tr_noise_stability}, with translations on the left and boundary pixels corruption on the right. 

In \autoref{sec:evidence_images} we then argue that differences in performance between the two training regimes can be explained by gaps in sensitivities with respect to input transformations that do not change the label. 
For \textit{(i)}, the gap is similar to the one observed for diffeomorphisms (\autoref{fig:stability_vs_n}). Still, the space of translations has negligible size with respect to input space, hence we expect the diffeomorphisms to have a more prominent effect. In case \textit{(ii)}, the feature regime is less sensitive with respect to irrelevant pixels corruption and this would give it an advantage over the lazy regime. The fact that the performance difference is in favor of the lazy regime instead, means that these transformations only play a minor role.

\newpage
\section{Maximum-entropy model of diffeomorphisms}
\label{app:diffeo}

We briefly review here the maximum-entropy model of diffeomorphisms as introduced in \cite{petrini_relative_2021}.

An image can be thought of as a function $x(s)$ describing intensity in position $s=(u,v)\in[0,1]^2 $, where $u$ and $v$ are the horizontal and vertical (pixel) coordinates. Denote $\tau x$ the image deformed by $\tau$, i.e. $[\tau x](s)=x(s-\tau(s))$. \cite{petrini_relative_2021} propose an ensemble of diffeomorphisms $\tau(s) = (\tau_u,\tau_v)$ with i.i.d. $\tau_u$ and $\tau_v$ defined as
\begin{equation}
    \tau_{u}=\sum_{i,j\in \mathbb{N}^+} C_{ij} \sin(i \pi u) \sin(j \pi v)
\end{equation}
 where the $C_{ij}$'s are Gaussian variables of zero mean and variance $T/(i^2+j^2)$ and $T$ is a parameter controlling the deformation magnitude. 
Once $\tau$ is generated, pixels are displaced to random positions. See \autoref{fig:diffeo_dog} for an example of such transformation.